\documentclass[journal]{IEEEtran}

\usepackage{ifpdf}
\usepackage{cite}
\usepackage[acronym, nomain, nopostdot]{glossaries}

\usepackage{import}
\usepackage{tikz, pgfplots, pgfkeys}
\usetikzlibrary{plotmarks, shapes, arrows, arrows.meta, shadows}
\usepackage{grffile}

\usepackage{color}
\usepackage{graphicx}

\usepackage{booktabs}

\usepackage{amsmath, amssymb}
\usepackage[noend]{algpseudocode}
\usepackage{algorithm}
\usepackage{setspace}
\usepackage{array}

\ifCLASSOPTIONcompsoc
  \usepackage[caption=false,font=normalsize,labelfont=sf,textfont=sf]{subfig}
\else
  \usepackage[caption=false,font=footnotesize]{subfig}
\fi

\usepackage{dblfloatfix}

\ifCLASSOPTIONcaptionsoff
  \usepackage[nomarkers]{endfloat}
 \let\MYoriglatexcaption\caption
 \renewcommand{\caption}[2][\relax]{\MYoriglatexcaption[#2]{#2}}
\fi

\usepackage{url}

\newacronym{ANOVA}{ANOVA}{analysis of variance}
\newacronym{ASA}{ASA}{auditory scene analysis}
\newacronym{ASM}{ASM}{acoustic scene mapping}
\newacronym{ASR}{ASR}{automatic speech recognition}
\newacronym{CMA-ES}{CMA-ES}{covariance matrix adaptation evolution strategy}
\newacronym{DLDS}{DLDS}{distributed linear dynamical system}
\newacronym{DDS}{DDS}{distributed dynamical system}
\newacronym{DKF}{DKF}{distributed Kalman filter}
\newacronym{DNN}{DNN}{deep neural network}
\newacronym{DoA}{DoA}{direction-of-arrival}
\newacronym{DSW}{DSW}{dynamic stream weight}
\newacronym{DPD}{DPD}{direct-path dominance}
\newacronym{DS}{DS}{dynamical system}
\newacronym{EKF}{EKF}{extended Kalman filter}
\newacronym{ES}{ES}{evolution strategies}
\newacronym{FPS}{FPS}{frames per second}
\newacronym{GM}{GM}{graphical model}
\newacronym{HMM}{HMM}{hidden Markov model}
\newacronym{IBF}{IBF}{information-based feedback}
\newacronym{IDP}{IDP}{inverse depth parametrization}
\newacronym{KAVLoC}{KAVLoC}{Kinect Audiovisual Localization Corpus}
\newacronym{KF}{KF}{Kalman filter}
\newacronym{KLD}{KLD}{Kullback-Leibler divergence}
\newacronym{LDS}{LDS}{linear dynamical system}
\newacronym{NES}{NES}{natural evolution strategies}
\newacronym{MCE}{MCE}{Monte Carlo exploration}
\newacronym{MDF}{MDF}{multisensor data fusion}
\newacronym{MCTS}{MCTS}{Monte Carlo tree search}
\newacronym{ML}{ML}{maximum likelihood}
\newacronym{MMSE}{MMSE}{minimum mean squared error}
\newacronym{MVAD}{MVAD}{multimodal voice activity detection}
\newacronym{MUSIC}{MUSIC}{multiple signal classification}
\newacronym{NAVLoC}{NAVLoC}{Nao Audiovisual Localization Corpus}
\newacronym{ODSW}{ODSW}{oracle dynamic stream weight}
\newacronym{PDF}{PDF}{probability density function}
\newacronym{PF}{PF}{particle filter}
\newacronym{PHD}{PHD}{probability hypothesis density}
\newacronym{RL}{RL}{reinforcement learning}
\newacronym{RMSE}{RMSE}{root mean square error}
\newacronym{ReLU}{ReLU}{rectified linear unit}
\newacronym{RVM}{RVM}{rotating vector model}
\newacronym{SGD}{SGD}{stochastic gradient descent}
\newacronym{SLAM}{SLAM}{simultaneous localization and mapping}
\newacronym{SNES}{SNES}{separable natural evolution strategies}
\newacronym{SNR}{SNR}{signal-to-noise ratio}
\newacronym{SRP-PHAT}{SRP-PHAT}{steered response power phase transform}
\newacronym{TDOA}{TDOA}{time-difference of arrival}
\newacronym{UKF}{UKF}{unscented Kalman filter}
\newacronym{VAD}{VAD}{voice activity detection}
\newacronym{WSN}{WSN}{wireless sensor network}

\begin{document}

\title{Audiovisual Speaker Tracking using Nonlinear Dynamical Systems with Dynamic Stream Weights}

\author{Christopher~Schymura
        and~Dorothea~Kolossa
\thanks{C. Schymura and D. Kolossa  are with the Cognitive Signal Processing Group, Institute of Communication Acoustics, Faculty of Electrical Engineering and Information Technology, Ruhr University Bochum, 44801 Bochum, Germany (e-mail: christopher.schymura@rub.de; dorothea.kolossa@rub.de).}}


\maketitle

\begin{abstract}
Data fusion plays an important role in many technical applications that require efficient processing of multimodal sensory observations. A prominent example is audiovisual signal processing, which has gained increasing attention in automatic speech recognition, speaker localization and related tasks. If appropriately combined with acoustic information, additional visual cues can help to improve the performance in these applications, especially under adverse acoustic conditions. A dynamic weighting of acoustic and visual streams based on instantaneous sensor reliability measures is an efficient approach to data fusion in this context. This paper presents a framework that extends the well-established theory of nonlinear dynamical systems with the notion of dynamic stream weights for an arbitrary number of sensory observations. It comprises a recursive state estimator based on the Gaussian filtering paradigm, which incorporates dynamic stream weights into a framework closely related to the extended Kalman filter. Additionally, a convex optimization approach to estimate oracle dynamic stream weights in fully observed dynamical systems utilizing a Dirichlet prior is presented. This serves as a basis for a generic parameter learning framework of dynamic stream weight estimators. The proposed system is application-independent and can be easily adapted to specific tasks and requirements. A study using audiovisual speaker tracking tasks is considered as an exemplary application in this work. An improved tracking performance of the dynamic stream weight-based estimation framework over state-of-the-art methods is demonstrated in the experiments.
\end{abstract}


\ifCLASSOPTIONpeerreview
 \begin{center} \bfseries EDICS Category: 3-BBND \end{center}
\fi
%
\IEEEpeerreviewmaketitle

\section{Introduction}
\label{sec:introduction}
\IEEEPARstart{E}{ffective} fusion of signals acquired from different sensory modalities is an important aspect of many technical applications. With the advent of emerging technologies like autonomous driving~\cite{Hossein2010, Ravanbakhsh2018}, assistive robotics~\cite{Katyal2013, Ivorra2018}, smart home environments~\cite{Fleury2010, Medjahed2011, Hossain2016} and \gls{ASR}~\cite{Potamianos2009, Petridis2018, Abdelaziz2018}, significant progress on the development of algorithms for \gls{MDF} has been made~\cite{Khaleghi2013}. The fundamental problem that \gls{MDF} tries to solve is, how signals obtained from different sensors can be combined to maximize information gain for the variables of interest~\cite{Zhang2006}, e.g. the location of a speaker or the transcription of a spoken sentence.

Many successful \gls{MDF} algorithms belong to the class of probabilistic fusion methods, often also denoted as Bayesian fusion. An exceptional example in this regard is the well-established \gls{KF}~\cite{Kalman1960}. Even though it relies on a linear Gaussian model, the \gls{KF} has gained tremendous success in a wide range of applications, due to its mathematical intuitiveness and computational efficiency~\cite{Khaleghi2013, Fang2018}. Since it was first introduced, many extensions to the \gls{KF} have been proposed, most notably the \gls{EKF}~\cite{McElhoe1966} and the \gls{UKF}~\cite{Julier1997}, which both overcome the linearity constraint. Additionally, \glspl{PF} were introduced as a framework to cope with nonlinear systems affected by non-Gaussian noise~\cite{Arulampalam2002}.

A common property of the aforementioned Bayesian fusion techniques is that they handle the actual sensor fusion implicitly. For instance, the standard \gls{KF} maintains a joint observation noise covariance matrix describing the noise characteristics of all sensors. This is an efficient approach for \gls{MDF}, as long as the reliabilities of individual sensors do not change over time. To overcome this constraint, adaptive \glspl{KF} have been proposed~\cite{Mehra1970}, which maintain and update an estimate of the observation noise covariance matrix at each time step. However, this adaptation process requires a sufficient amount of past observations to achieve reliable estimates of the observation noise covariance matrix. An optimal adaptation framework would allow to weight the individual contributions of each sensor instantaneously.

This was recently proposed in~\cite{Schymura2018}, which presented an inference scheme for continuous state spaces in the regime of linear \glspl{DS} that incorporated \glspl{DSW} into the estimation process. \glspl{DSW} are weighting factors that control the contribution of the individual sensory observations to the estimation process at each time-step. The work was inspired by previous applications of \glspl{DSW}, which were initially proposed in the context of audiovisual \gls{ASR}. Pioneering work in this regard was conducted in~\cite{Abdelaziz2015} for probabilistic inference incorporating \glspl{DSW} into \glspl{HMM}. Compared to conventional Bayesian fusion techniques, this allows to rapidly adapt the estimation process. However, predicting \glspl{DSW} requires the availability of instantaneous sensor reliability measures, which depend on sensor type and the desired application. Research on appropriate reliability measures has already been conducted in the context of audiovisual \gls{ASR}~\cite{Gergen2016, Meutzner2017}. A standard approach is the computation of \glspl{ODSW} using training data with available ground-truth information and subsequently perform supervised training, using e.g. a regression function or a neural network~\cite{Abdelaziz2015, Meutzner2017, Abdelaziz2018}.

This paper introduces an extended framework for \gls{MDF} using nonlinear \glspl{DS} based on the method initially proposed in~\cite{Schymura2018}. This method was restricted to a linear Gaussian state space representation and two sensory modalities. A general recursive inference method for state estimation that can be applied to nonlinear \glspl{DS} with Gaussian noise, \glspl{DSW} and an arbitrary number of observations is proposed in this paper. Furthermore, a means to obtain \glspl{ODSW} from fully observed \glspl{DS} based on a Dirichlet prior imposed on the stream weights is presented. Compared to the previously introduced \gls{ODSW} estimators using a Gaussian prior~\cite{Abdelaziz2015, Schymura2018}, this novel approach provides a clear and intuitive probabilistic interpretation. Additionally, a generic learning scheme for \glspl{DSW} prediction models using application-specific reliability measures is presented. It allows to train a broad class of models, whose sole restrictions are differentiability with respect to the function parameters and a softmax output function. Hence, nonlinear models like e.g. \glspl{DNN} are naturally supported as potential \gls{DSW} estimators. 

The application of audiovisual speaker tracking is considered in this study. It is well suited for evaluating Bayesian \gls{MDF} approaches with continuous state spaces, as the variables of interest are encoded as either Cartesian coordinates or \gls{DoA} values. Furthermore, speaker localization scenarios involve highly dynamic components if speakers are moving within the environment and have to cope with various types of disturbances. This includes background noise and reverberation affecting the acoustic signals, as well as changing lighting conditions and occlusion affecting the video signals. It should be noted that the entire framework proposed in this paper is not restricted to a particular application, but should rather be considered as a generic approach to Bayesian \gls{MDF} incorporating \glspl{DSW}.

A variety of related models for Bayesian fusion have been previously proposed for many technical applications. To put the work presented in this study into context, related existing approaches will be briefly reviewed in the following.

Early work on \gls{MDF} introduced a special class of \glspl{DS} that utilize standard \glspl{KF} and their extensions for Bayesian fusion. These systems are referred to as \glspl{DDS}~\cite{Speyer1979, Rao1993, Antonelli2013}. They provide a natural extension to the \gls{DS} paradigm by incorporating multiple independent sensors with distinct observation models and noise characteristics. Prominent application domains for \glspl{DDS} are wireless sensor networks~\cite{Olfati-Saber2005, Solouk2013} and multiagent systems~\cite{Yu2010}. The mathematical foundations of \glspl{DDS} provide a generic framework for modeling systems with multimodal sensory input. For instance, the work reported in~\cite{Olfati-Saber2005} proposes a distributed \gls{KF} for state estimation in wireless sensor networks. It is based on decomposing the standard \gls{KF} into a set of so-called micro \glspl{KF} for each of the individual sensory observations. This results in a network of \glspl{KF} which is capable of collectively estimating the system state. A theoretical analysis of similar inference algorithms for \glspl{DDS} is given in~\cite{Funiak2007}. 

Besides the many contributions in the field of wireless sensor networks, further successful approaches to Bayesian \gls{MDF} have been proposed in other technical fields. Focusing on the domain of audiovisual signal processing, a variety of algorithms for audiovisual speaker tracking is available. For instance, the framework described in~\cite{Gehrig2005} uses an \gls{EKF} with a joint audiovisual observation vector to localize and track speakers during recorded seminars. It does not incorporate a distributed architecture of the underlying \gls{DS}, but rather handles data fusion implicitly during the recursive update step. Another approach was introduced in~\cite{Gerlach2012}, where a \gls{PF} was utilized to localize and track speakers in domestic environments. The framework provided explicit control over the individual contributions of acoustic and visual observations via exponential weighting parameters, which were determined a-priori using a grid-search. A recently proposed algorithm for speaker tracking has explicitly considered sensor reliability measures within a particle filtering framework~\cite{Qian2017}. This work utilized the peak value of the acoustic global coherence field and the correlation between a color-histogram template and the detected face as features, which affected the weighting and resampling step of the particle filter. 

A noteworthy study that fits nicely into the context of this work is the framework based on distributed multi-sensor, multi-target tracking presented in~\cite{Uney2013}. It proposes a recursive Bayesian filter that assigns weights to sensory observations based on exponential mixture densities. This representation of the filtering distribution is mathematically similar to the framework proposed in this paper. However, the weighting scheme serves a different purpose, namely, optimizing track-to-track fusion in multi-object distributions. This stands in contrast to the present study, where the weighting is applied instantaneously rather than using fixed weighting factors.

In this regard, the framework presented in this paper is most closely related to approaches developed for discrete state spaces based on \glspl{HMM} in audiovisual \gls{ASR}~\cite{Abdelaziz2015, Meutzner2017, Gergen2016, Abdelaziz2018}. The primary contribution of the present work is the introduction of \gls{DSW}-based \gls{MDF} into continuous \glspl{DS} with nonlinear dynamics and observations, which is a natural extension of the initial work reported in~\cite{Schymura2018}. Additionally, a novel approach for computing \glspl{ODSW} incorporating a Dirichlet prior is derived and integrated into a generic learning framework that allows to train \gls{DSW} prediction models.

\section{Dynamical System Description}
\label{sec:dynamical_system_description}
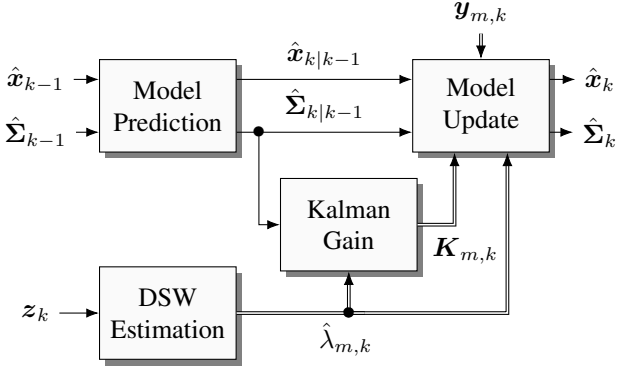
\begin{figure}[t]
	\centering
	\pgfdeclarelayer{background}
\pgfdeclarelayer{foreground}
\pgfsetlayers{background,main,foreground}

\definecolor{rub_green}{rgb}{0.5804,0.7569,0.1059}%
\definecolor{rub_blue}{rgb}{0,0.2078,0.3765}%
\definecolor{rub_gray}{rgb}{0.8666,0.8666,0.8666}%
\definecolor{rub_orange}{rgb}{0.8941,0.5333,0.2549}

\tikzstyle{block} = [draw, text width=4.5em, text centered, fill=black!02, minimum height=3.5em, drop shadow=black]



\begin{tikzpicture}
	
	\node (model_prediction) [block] {Model Prediction};
	\node (model_update) [block, right of=model_prediction, xshift=9em] {Model Update};
	\node (stream_weight_prediction) [block, below of=model_prediction, yshift=-5em] {DSW Estimation};
	\node (kalman_gain) [block, right of=model_prediction, xshift=4em, yshift=-4.5em] {Kalman Gain};
	
	\node (x_hat_prev) [left of=model_prediction, node distance = 5em, yshift=1em] {$\hat{\boldsymbol{x}}_{k - 1}$};
	\node (sigma_hat_prev) [left of=model_prediction, node distance = 5em, yshift=-1em] {$\hat{\boldsymbol{\Sigma}}_{k - 1}$};
	
	\node (y) [above of=model_update, node distance = 3.5em] {$\boldsymbol{y}_{m, k}$};
	
	\draw [double,-{Latex[scale=0.9]}] (y.south) -| node[auto] {} (model_update.north);
	
	\node (z) [left of=stream_weight_prediction, node distance = 5em] {$\boldsymbol{z}_{k}$};
	
	\node (x_hat_post) [right of=model_update, node distance = 4.5em, yshift=1em] {$\hat{\boldsymbol{x}}_{k}$};
	\node (sigma_hat_post) [right of=model_update, node distance = 4.5em, yshift=-1em] {$\hat{\boldsymbol{\Sigma}}_{k}$};	
	
	\draw [-{Latex[scale=1.0]}] (x_hat_prev) -- node[auto] {} ([yshift=1em]model_prediction.west);
	\draw [-{Latex[scale=1.0]}] (sigma_hat_prev) -- node[auto] {} ([yshift=-1em]model_prediction.west);
	
	\draw [-{Latex[scale=1.0]}] (z) -- node[auto] {} (stream_weight_prediction.west);
	
	\draw [-{Latex[scale=1.0]}] ([yshift=1em]model_prediction.east) -- node[auto] {$\hat{\boldsymbol{x}}_{k | k - 1}$} ([yshift=1em]model_update.west);
	\draw [-{Latex[scale=1.0]}] ([yshift=-1em]model_prediction.east) -- node[auto] {$\hat{\boldsymbol{\Sigma}}_{k | k - 1}$} ([yshift=-1em]model_update.west);
	
	\draw [-{Latex[scale=1.0]}] ([yshift=1em]model_update.east) -- node[auto] {} (x_hat_post);
	\draw [-{Latex[scale=1.0]}] ([yshift=-1em]model_update.east) -- node[auto] {} (sigma_hat_post);
	
	\node (intersection_01) [right of=model_prediction,xshift=0.6em,yshift=-1em] {\textbullet};
	\node (intersection_02) [right of=stream_weight_prediction,xshift=4em] {\textbullet};
	
	\draw [-{Latex[scale=1.0]}] ([yshift=0.75em]intersection_01.south) |- node[auto] {} (kalman_gain.west);
	
	\draw [double,-{Latex[scale=0.9]}] (kalman_gain.east) -| node[xshift=2em,yshift=-1em,auto] {$\boldsymbol{K}_{m, k}$} ([xshift=-1em]model_update.south);
	
	\draw [double,-{Latex[scale=0.9]}] (stream_weight_prediction.east) -| node[xshift=-4.75em,yshift=-1em,auto] {$\hat{\lambda}_{m, k}$} ([xshift=1em]model_update.south);
	
	\draw [double,-{Latex[scale=0.9]}] (intersection_02) -| node[auto] {} (kalman_gain.south);
	
	\node[draw=none,fill=none] at (intersection_02) (intersection_02_bullet) {\textbullet};

\end{tikzpicture}%
	\caption{Block diagram of the proposed state estimation framework. Double arrows indicate the simultaneous data transfer of multiple variables. The filter recursively updates estimates of the state mean \(\hat{\boldsymbol{x}}_{k}\) and covariance matrix \(\hat{\boldsymbol{\Sigma}}_{k}\) by incorporating observations \(\boldsymbol{y}_{m, k}\) from \(m = 1,\,\ldots,\,M\) sensors and predicted \glspl{DSW} \(\hat{\lambda}_{m, k}\). A core aspect of this recursive update is the computation of the Kalman gains \(\boldsymbol{K}_{m, k}\). The \glspl{DSW} are estimated from reliability measures \(\boldsymbol{z}_{k}\) by a prediction model obtained via supervised training.}
	\label{fig:block_diagram}
\end{figure}

This section presents an extension of the state estimation framework proposed in~\cite{Schymura2018} to nonlinear \glspl{DS} with \glspl{DSW} and an arbitrary number of independent observations. A generic state estimation algorithm based on the Gaussian filter paradigm is derived and its relation to the standard \gls{EKF} is discussed. A structural overview of the proposed system is depicted in Fig.~\ref{fig:block_diagram}. It illustrates the relation between all system components that will be described in the following sections.

\subsection{Nonlinear system model}
\label{subsec:nonlinear_system_model}
Consider an autonomous, discrete-time nonlinear \gls{DS} with Gaussian noise and \(M\) independent observations
\begin{align}
\boldsymbol{x}_{k} &= f(\boldsymbol{x}_{k - 1}) + \boldsymbol{v}_{k}, \label{eqn:lds_process_model} \\
\boldsymbol{y}_{m, k} &= h_{m}(\boldsymbol{x}_{k}) + \boldsymbol{w}_{m, k}, \label{eqn:lds_observation_model}
\end{align}
where \(\boldsymbol{x}_{k} \in \mathbb{R}^{D_{x}}\) denotes the state vector at discrete time step \(k\) and \(\boldsymbol{y}_{m, k}\) represents the \(m\)-th observation vector with \(m = 1,\,\ldots,\,M\). The system dynamics are governed by the state transition function \(f(\boldsymbol{x}_{k - 1})\) and zero-mean Gaussian noise \(\boldsymbol{v}_{k} \sim \mathcal{N}(\boldsymbol{0},\,\boldsymbol{Q})\) with covariance matrix \(\boldsymbol{Q} \in \mathbb{R}^{D_{x} \times D_{x}}\). The state-to-observation transformations are described by \(M\) observation functions \(h_{m}(\boldsymbol{x}_{k})\), which are affected by zero-mean Gaussian noise terms \(\boldsymbol{w}_{m, k} \sim \mathcal{N}(\boldsymbol{0},\,\boldsymbol{R}_{m})\) with covariance matrices \(\boldsymbol{R}_{m} \in \mathbb{R}^{D_{y, m} \times D_{y, m}}\). Autonomous \glspl{DS} are considered in this work as they are widely used in localization and tracking applications. However, an extension of the proposed methods to \glspl{DS} with external input is generally possible.

Following the approach proposed in~\cite{Schymura2018}, the incorporation of \glspl{DSW} \(\lambda_{m, k}\) allows to express the joint likelihood function of the \gls{DS} described by Eqs.~\eqref{eqn:lds_process_model}--\eqref{eqn:lds_observation_model} up to time step \(k\) as
\begin{align}
p(\mathcal{X}_{0 : k}&,\,\mathcal{Y}_{1, 1 : k},\,\ldots,\,\mathcal{Y}_{M, 1 : k}) \propto \notag \\
& p(\boldsymbol{x}_{0}) \prod_{k^{\prime} = 1}^{k} p(\boldsymbol{x}_{k^{\prime}}\,|\,\boldsymbol{x}_{k^{\prime} - 1}) \prod_{m = 1}^{M} p(\boldsymbol{y}_{m, k^{\prime}}\,|\,\boldsymbol{x}_{k^{\prime}})^{\lambda_{m, k^{\prime}}}, \label{eqn:lds_joint_likelihood}
\end{align}
where \(\mathcal{X}_{0 : k} = \{\boldsymbol{x}_{0},\,\ldots,\boldsymbol{x}_{k}\}\) and \(\mathcal{Y}_{m, 1 : k} = \{\boldsymbol{y}_{m, 1},\,\ldots,\boldsymbol{y}_{m, k}\}\) are the corresponding sequences (also referred to as trajectories) of state and observation vectors. The \glspl{DSW} in Eq.~\eqref{eqn:lds_joint_likelihood} must satisfy the constraint \(\sum_{m = 1}^{M} \lambda_{m, k} = 1~\forall\,k\). 
\subsection{State estimation}
\label{subsec:state_estimation}
A Gaussian filter to infer the state of the \gls{DS} can be derived by marginalizing out the previous states in the joint likelihood function in Eq.~\eqref{eqn:lds_joint_likelihood}. This yields the well-known prediction and update steps of the Bayes filter~\cite[Chap.~2]{Thrun2005}, given by
\begin{align}
&p(\boldsymbol{x}_{k}\,|\,\mathcal{Y}_{1, 1 : k - 1},\,\ldots,\,\mathcal{Y}_{M, 1 : k - 1}) = \notag \\
&\hspace{-0.11cm}\int p(\boldsymbol{x}_{k}\,|\,\boldsymbol{x}_{k - 1}) p(\boldsymbol{x}_{k - 1}\,|\,\mathcal{Y}_{1, 1 : k - 1},\,\ldots,\,\mathcal{Y}_{M, 1 : k - 1})\,d\boldsymbol{x}_{k - 1} \label{eqn:app:prediction_step}
\end{align}
and
\begin{align}
p&(\boldsymbol{x}_{k}\,|\,\mathcal{Y}_{1, 1 : k},\,\ldots,\,\mathcal{Y}_{M, 1 : k}) \propto \notag \\
&\,p(\boldsymbol{x}_{k}\,|\,\mathcal{Y}_{1, 1 : k - 1},\,\ldots,\,\mathcal{Y}_{M, 1 : k - 1}) \prod_{m = 1}^{M} p(\boldsymbol{y}_{m, k}\,|\,\boldsymbol{x}_{k})^{\lambda_{m, k}}. \label{eqn:app:update_step}
\end{align}
Assuming that the first derivatives of the state transition function and the observation functions in Eqs.~\eqref{eqn:lds_process_model}--\eqref{eqn:lds_observation_model} exist, a first-order Taylor series expansion about the estimated state posterior mean \(\hat{\boldsymbol{x}}_{k} = E\{\boldsymbol{x}_{k}\,|\,\mathcal{Y}_{1, 1 : k},\,\ldots,\,\mathcal{Y}_{M, 1 : k}\}\) can be expressed as
\begin{equation}
f(\boldsymbol{x}_{k - 1}) \approx f(\hat{\boldsymbol{x}}_{k - 1}) + \boldsymbol{F}(\hat{\boldsymbol{x}}_{k - 1}) \boldsymbol{\delta}_{k - 1}
\label{eqn:app:taylor_state_transition}
\end{equation}
and
\begin{equation}
h_{m}(\boldsymbol{x}_{k}) \approx h_{m}(\hat{\boldsymbol{x}}_{k}) + \boldsymbol{H}_{m}(\hat{\boldsymbol{x}}_{k}) \boldsymbol{\delta}_{k},
\label{eqn:app:taylor_observation}
\end{equation}
with \(\boldsymbol{\delta}_{k} = \boldsymbol{x}_{k} - \hat{\boldsymbol{x}}_{k}\), where \(\boldsymbol{F}(\hat{\boldsymbol{x}}_{k - 1}) \in \mathbb{R}^{D_{x} \times D_{x}}\) is the Jacobian of the state transition function and \(\boldsymbol{H}_{m}(\hat{\boldsymbol{x}}_{k}) \in \mathbb{R}^{D_{y, m} \times D_{x}}\) is the Jacobian of the \(m\)-th observation function, respectively. This approach is equivalent to the derivation of the \gls{EKF}~\cite{McElhoe1966}. For notational convenience, the explicit dependency of the Jacobians on the state will be omitted  in the following sections, according to \(\boldsymbol{F}(\hat{\boldsymbol{x}}_{k - 1}) \equiv \boldsymbol{F}_{k - 1}\) and \(\boldsymbol{H}_{m}(\hat{\boldsymbol{x}}_{k}) \equiv \boldsymbol{H}_{m, k}\). This allows to express the \glspl{PDF} in Eqs.~\eqref{eqn:app:prediction_step} and~\eqref{eqn:app:update_step} as
\begin{align}
p(\boldsymbol{x}_{k}\,|\,\boldsymbol{x}_{k - 1}) &= \mathcal{N}\Big(\boldsymbol{x}_{k}\,|\,f(\hat{\boldsymbol{x}}_{k - 1}) + \boldsymbol{F}_{k - 1}\boldsymbol{\delta}_{k - 1},\,\boldsymbol{Q}\Big), \label{eqn:app:state_pdf} \\
p(\boldsymbol{y}_{m, k}\,|\,\boldsymbol{x}_{k}) &= \mathcal{N}\Big(\boldsymbol{y}_{m, k}\,|\,h_{m}(\hat{\boldsymbol{x}}_{k}) + \boldsymbol{H}_{m, k} \boldsymbol{\delta}_{k},\,\boldsymbol{R}_{m}\Big) \label{eqn:app:observation_pdf}
\end{align}
and
\begin{equation}
p(\boldsymbol{x}_{k}\,|\,\mathcal{Y}_{1, 1 : k},\,\ldots,\,\mathcal{Y}_{M, 1 : k}) = \mathcal{N}\Big(\boldsymbol{x}_{k}\,|\,\hat{\boldsymbol{x}}_{k},\,\hat{\boldsymbol{\Sigma}}_{k}\Big),
\label{eqn:app:recursive_state_pdf}
\end{equation}
where \(\hat{\boldsymbol{\Sigma}}_{k}\) is the estimated state posterior covariance matrix, which needs to be updated conjointly with the estimated state posterior mean \(\hat{\boldsymbol{x}}_{k}\) at each time step. This update will be performed recursively via the prediction and update steps of the Gaussian filter.

The prediction step is obtained by inserting Eqs.~\eqref{eqn:app:state_pdf} and~\eqref{eqn:app:recursive_state_pdf} into Eq.~\eqref{eqn:app:prediction_step}, taking the first and second derivative and solving for the predicted state mean \(\hat{\boldsymbol{x}}_{k | k - 1}\) and the predicted state covariance matrix \(\hat{\boldsymbol{\Sigma}}_{k | k - 1}\). The resulting equations
given in Alg.~\ref{alg:lds_inference} are identical to the prediction step of the \gls{EKF}. Hence, the derivation is omitted here, cf.~\cite[Chap.~3]{Thrun2005} for details.

For the derivation of the update step, Eqs.~\eqref{eqn:app:observation_pdf} and~\eqref{eqn:app:recursive_state_pdf} are inserted into the the log-likelihood form of Eq.~\eqref{eqn:app:update_step}, yielding
\begin{align}
\log\Big\{p&(\boldsymbol{x}_{k}\,|\,\mathcal{Y}_{1, 1 : k},\,\ldots,\,\mathcal{Y}_{M, 1 : k})\Big\} = \mathrm{const.} + \notag \\
& \quad~ (\boldsymbol{x}_{k} - \hat{\boldsymbol{x}}_{k | k - 1})^{\mathrm{T}} \hat{\boldsymbol{\Sigma}}_{k | k - 1}^{-1} (\boldsymbol{x}_{k} - \hat{\boldsymbol{x}}_{k | k - 1}) \notag \\
& + \sum_{m = 1}^{M} \lambda_{m, k} \Big[\Big(\boldsymbol{y}_{m, k} - h_{m}(\hat{\boldsymbol{x}}_{k}) - \boldsymbol{H}_{m, k}(\boldsymbol{x}_{k} - \hat{\boldsymbol{x}}_{k})\Big)^{\mathrm{T}} \notag \\
& \times \boldsymbol{R}_{m}^{-1} \Big(\boldsymbol{y}_{m, k} - h_{m}(\hat{\boldsymbol{x}}_{k}) - \boldsymbol{H}_{m, k}(\boldsymbol{x}_{k} - \hat{\boldsymbol{x}}_{k})\Big)\Big]. \label{eqn:update_log_likelihood}
\end{align}
Taking the first and second derivative of Eq.~\eqref{eqn:update_log_likelihood}, where the system state \(\boldsymbol{x}_{k}\) has been substituted with the estimated state posterior mean \(\hat{\boldsymbol{x}}_{k}\), results in the expressions
\begin{align}
\frac{\partial}{\partial \boldsymbol{x}_{k}}&\log\Big\{p(\boldsymbol{x}_{k}\,|\,\mathcal{Y}_{1, 1 : k},\,\ldots,\,\mathcal{Y}_{M, 1 : k})\Big\}\Big|_{\boldsymbol{x}_{k} = \hat{\boldsymbol{x}}_{k}} = \notag \\
& \quad~ \hat{\boldsymbol{\Sigma}}_{k | k - 1}^{-1} (\hat{\boldsymbol{x}}_{k} - \hat{\boldsymbol{x}}_{k | k - 1}) \notag \\
&+ \sum_{m = 1}^{M} \lambda_{m, k} \Big[ \boldsymbol{H}_{m, k}^{\mathrm{T}} \boldsymbol{R}_{m}^{-1} \Big(h_{m}(\hat{\boldsymbol{x}}_{k | k - 1}) - \boldsymbol{y}_{m, k}\Big) \notag \\
&+ \boldsymbol{H}_{m, k}^{\mathrm{T}} \boldsymbol{R}_{m}^{-1} \boldsymbol{H}_{m, k} \Big(\hat{\boldsymbol{x}}_{k} - \hat{\boldsymbol{x}}_{k | k - 1}\Big)\Big] \label{eqn:update_log_likelihood_first_derivative}
\end{align}
and
\begin{align}
\frac{\partial^{2}}{\partial \boldsymbol{x}_{k}^{2}}\log&\Big\{p(\boldsymbol{x}_{k}\,|\,\mathcal{Y}_{1, 1 : k},\,\ldots,\,\mathcal{Y}_{M, 1 : k})\Big\}\Big|_{\boldsymbol{x}_{k} = \hat{\boldsymbol{x}}_{k}} = \notag \\
& \hat{\boldsymbol{\Sigma}}_{k | k - 1}^{-1} + \sum_{m = 1}^{M} \lambda_{m, k} \Big[\boldsymbol{H}_{m, k}^{\mathrm{T}} \boldsymbol{R}_{m}^{-1} \boldsymbol{H}_{m, k}\Big]. \label{eqn:update_log_likelihood_second_derivative}
\end{align}
The second derivative in Eq.~\eqref{eqn:update_log_likelihood_second_derivative} is the curvature of the quadratic function in Eq.~\eqref{eqn:update_log_likelihood}, whose inverse is the covariance matrix of the state posterior \(p(\boldsymbol{x}_{k}\,|\,\mathcal{Y}_{1, 1 : k},\,\ldots,\,\mathcal{Y}_{M, 1 : k})\), cf.~\cite[Chap.~3]{Thrun2005}. Therefore, a closed-form expression
\begin{equation}
\hat{\boldsymbol{\Sigma}}_{k} = \Big(\boldsymbol{I} - \sum_{m = 1}^{M} \lambda_{m, k} \boldsymbol{K}_{m, k} \boldsymbol{H}_{m, k}\Big) \hat{\boldsymbol{\Sigma}}_{k | k - 1}
\label{eqn:state_posterior_covariance}
\end{equation}
for the estimated state posterior covariance matrix can be obtained, where
\begin{equation}
\boldsymbol{K}_{m, k} = \hat{\boldsymbol{\Sigma}}_{k} \boldsymbol{H}_{m, k}^{\mathrm{T}} \boldsymbol{R}_{m}^{-1}
\label{eqn:kalman_gain}
\end{equation}
\begin{figure*}[!t]
	\setstretch{1.25}
	\normalsize	
	\begin{equation}
	\begin{bmatrix}
	\boldsymbol{R}_{1} + \lambda_{1, k} \boldsymbol{H}_{1, k} \hat{\boldsymbol{\Sigma}}_{k | k - 1} \boldsymbol{H}_{1, k}^{\mathrm{T}} & \cdots & \lambda_{M, k} \boldsymbol{H}_{1, k} \hat{\boldsymbol{\Sigma}}_{k | k - 1} \boldsymbol{H}_{M, k}^{\mathrm{T}} \\
	\vdots & \ddots & \vdots \\
	\lambda_{1, k} \boldsymbol{H}_{M, k} \hat{\boldsymbol{\Sigma}}_{k | k - 1} \boldsymbol{H}_{1, k}^{\mathrm{T}} & \cdots & \boldsymbol{R}_{M} + \lambda_{M, k} \boldsymbol{H}_{M, k} \hat{\boldsymbol{\Sigma}}_{k | k - 1} \boldsymbol{H}_{M, k}^{\mathrm{T}}
	\end{bmatrix}\,\begin{bmatrix}
	\boldsymbol{K}_{1, k}^{\mathrm{T}} \\
	\vdots \\
	\boldsymbol{K}_{M, k}^{\mathrm{T}}
	\end{bmatrix} = \begin{bmatrix}
	\boldsymbol{H}_{1, k} \hat{\boldsymbol{\Sigma}}_{k | k - 1} \\
	\vdots \\
	\boldsymbol{H}_{M, k} \hat{\boldsymbol{\Sigma}}_{k | k - 1}
	\end{bmatrix}
	\label{eqn:kalman_gain_sle}
	\end{equation}	
	\hrulefill
\end{figure*}
\noindent is defined as the Kalman gain corresponding to the \(m\)-th observation. To resolve the dependency of Eq.~\eqref{eqn:kalman_gain} on the estimated state posterior covariance matrix \(\hat{\boldsymbol{\Sigma}}_{k}\), Eq.~\eqref{eqn:state_posterior_covariance} is inserted into Eq.~\eqref{eqn:kalman_gain}, which allows to derive an analytic solution for the individual Kalman gains \(\boldsymbol{K}_{m, k}\), by solving the system of linear matrix equations shown in Eq.~\eqref{eqn:kalman_gain_sle}. This system can be expressed as
\begin{equation}
\begin{bmatrix}
\boldsymbol{R} + \boldsymbol{U}_{k} \boldsymbol{W}_{k} \boldsymbol{U}_{k}^{\mathrm{T}}
\end{bmatrix} \boldsymbol{K}_{k} = \boldsymbol{B}_{k} \hat{\boldsymbol{\Sigma}}_{k | k - 1},
\label{eqn:sle}
\end{equation}
where \(\boldsymbol{R} = \mathrm{blkdiag}(\boldsymbol{R}_{1},\,\ldots,\,\boldsymbol{R}_{M})\) is a block-diagonal matrix composed of all corresponding observation noise covariance matrices, \(\boldsymbol{U}_{k} = \mathrm{blkdiag}(\boldsymbol{H}_{1, k},\,\ldots,\,\boldsymbol{H}_{M, k})\) comprises all observation Jacobians, \(\boldsymbol{W}_{k} = \boldsymbol{L}_{k} \otimes \hat{\boldsymbol{\Sigma}}_{k | k - 1}\) with
\begin{equation}
\boldsymbol{L}_{k} = \begin{bmatrix}
\lambda_{1, k} & \cdots & \lambda_{M, k} \\
\vdots & \ddots & \vdots \\
\lambda_{1, k} & \cdots & \lambda_{M, k}
\end{bmatrix}
\label{eqn:lambda_matrix}
\end{equation}
and \(\boldsymbol{B}_{k} = \begin{bmatrix}\boldsymbol{H}_{1, k} & \cdots & \boldsymbol{H}_{M, k}\end{bmatrix}^{\mathrm{T}}\). The Kalman gain solution matrix \(\boldsymbol{K} = \begin{bmatrix}\boldsymbol{K}_{1, k}^{\mathrm{T}} & \cdots & \boldsymbol{K}_{M, k}^{\mathrm{T}}\end{bmatrix}^{\mathrm{T}}\) contains all Kalman gains associated with the individual observations. A solution of Eq.~\eqref{eqn:sle} is obtained via the binomial inverse theorem~\cite{Harville1976}, where \(\boldsymbol{W}_{k}\) is always singular for \(M > 1\), which is shown in detail in App.~\ref{sec:app:singularity_of_w}. This yields
\begin{equation}
\boldsymbol{K}_{k} = \Big(\boldsymbol{R}^{-1} - \boldsymbol{R}^{-1} \boldsymbol{U}_{k} \boldsymbol{\Gamma}_{k} \boldsymbol{U}_{k}^{\mathrm{T}} \boldsymbol{R}^{-1}\Big) \boldsymbol{B}_{k} \hat{\boldsymbol{\Sigma}}_{k | k - 1},
\label{eqn:kalman_gain_solution}
\end{equation}
with \(\boldsymbol{\Gamma}_{k} = \boldsymbol{W}_{k} (\boldsymbol{I} + \boldsymbol{U}_{k}^{\mathrm{T}} \boldsymbol{R}^{-1} \boldsymbol{U}_{k} \boldsymbol{W}_{k})^{-1}\), which allows an efficient computation of the Kalman gains at each step, as the inverse of the observation noise block-diagonal covariance matrix can be precomputed.

The corresponding state update recursion are obtained by inserting Eq.~\eqref{eqn:kalman_gain} into Eq.~\eqref{eqn:update_log_likelihood_first_derivative}, exploiting the relationship in Eq.~\eqref{eqn:state_posterior_covariance} and solving for the estimated state posterior mean
\begin{equation}
\hat{\boldsymbol{x}}_{k} = \hat{\boldsymbol{x}}_{k | k - 1} + \sum_{m = 1}^{M} \lambda_{m, k} \boldsymbol{K}_{m, k} \Big(\boldsymbol{y}_{m, k} - h_{m}(\hat{\boldsymbol{x}}_{k | k - 1})\Big).
\label{eqn:state_posterior_mean}
\end{equation}

The resulting prediction and update steps of the presented Gaussian filtering algorithm are summarized in Alg.~\ref{alg:lds_inference}. An interactive Python implementation of the proposed algorithm is available online\footnote{\url{https://github.com/rub-ksv/avtrack}}.
\begin{algorithm}[t]
	\caption{Prediction and update steps of the Gaussian filtering framework for nonlinear dynamical systems with \(M\) independent observations and dynamic stream weights.}
	\begin{algorithmic}[1]
		\setstretch{1.33}
		
		\Function{Predict}{$\hat{\boldsymbol{x}}_{k - 1},\,\hat{\boldsymbol{\Sigma}}_{k - 1},\,\boldsymbol{Q}$}
		\State \vspace{-0.88cm}\begin{flalign*}
		\quad~\, \hat{\boldsymbol{x}}_{k | k - 1} = f(\hat{\boldsymbol{x}}_{k - 1}) &&
		\end{flalign*}
		\State \vspace{-0.85cm}\begin{flalign*}
		\quad~\, \hat{\boldsymbol{\Sigma}}_{k | k - 1} = \boldsymbol{F}_{k - 1} \hat{\boldsymbol{\Sigma}}_{k - 1} \boldsymbol{F}_{k - 1}^{\mathrm{T}} + \boldsymbol{Q} &&
		\end{flalign*}%
		\State \Return $\hat{\boldsymbol{x}}_{k | k - 1},\,\hat{\boldsymbol{\Sigma}}_{k | k - 1}$
		\EndFunction
		
		\setcounter{ALG@line}{0}
		
		\Statex
		
		\Function{Update}{$\hat{\boldsymbol{x}}_{k | k - 1},\,\hat{\boldsymbol{\Sigma}}_{k | k - 1},\,\boldsymbol{y}_{1, k},\,\ldots,\,\boldsymbol{y}_{M, k},\,\quad~\,$ $~~~~~\,\lambda_{1, k},\,\ldots,\,\lambda_{M, k},\,\boldsymbol{R}_{1},\,\ldots,\,\boldsymbol{R}_{M}$}		
		\State \vspace{-0.88cm}\begin{flalign*}
		\quad~\, \tilde{\boldsymbol{y}}_{m, k} = \boldsymbol{y}_{m, k} - h_{m}(\hat{\boldsymbol{x}}_{k | k - 1}) &&
		\end{flalign*}
		\State Compute $\boldsymbol{K}_{1, k},\,\ldots,\,\boldsymbol{K}_{M, k}$ using Eq.~\eqref{eqn:kalman_gain_solution}.\vspace{0.28cm}%
		\State \vspace{-0.95cm}\begin{flalign*}
		\quad~\, \hat{\boldsymbol{x}}_{k} = \hat{\boldsymbol{x}}_{k | k - 1} + \sum_{m = 1}^{M} \lambda_{m, k} \boldsymbol{K}_{m, k} \tilde{\boldsymbol{y}}_{m, k} &&
		\end{flalign*}\vspace{-0.3cm}%
		\State \vspace{-0.93cm}\begin{flalign*}
		\quad~\, \hat{\boldsymbol{\Sigma}}_{k} = \Big(\boldsymbol{I} - \sum_{m = 1}^{M} \lambda_{m, k} \boldsymbol{K}_{m, k} \boldsymbol{H}_{m, k}\Big) \hat{\boldsymbol{\Sigma}}_{k | k - 1} &&
		\end{flalign*}
		\State \Return $\hat{\boldsymbol{x}}_{k},\,\hat{\boldsymbol{\Sigma}}_{k}$		
		\EndFunction
	\end{algorithmic}
	\label{alg:lds_inference}
\end{algorithm}

\subsection{Comparison with the extended Kalman filter}
\label{subsec:ekf_comparison}
The state estimation framework presented in this work is a generalization of the standard \gls{EKF}, which is covered as a special case. This can be easily verified by evaluating Eqs.~\eqref{eqn:state_posterior_covariance},~\eqref{eqn:kalman_gain_sle} and~\eqref{eqn:state_posterior_mean} for \(M = 1\) and \(\lambda_{1, k} = 1~\forall\,k\), which yields the conventional \gls{EKF} update step. Both methods rely on a first-order Taylor expansion of the nonlinear state transition and observation functions. However, the standard \gls{EKF} is not capable of incorporating \glspl{DSW}, which is a unique property of the algorithm proposed here.

\section{Oracle dynamic stream weights}
\label{sec:odsw_dirichlet}
\glspl{ODSW} have already been thoroughly investigated in the context of \gls{ASR}, where they were utilized in \gls{HMM}-based recognizers with audiovisual input~\cite{Estellers2012, Abdelaziz2015, Gergen2016, Abdelaziz2018, Freiwald2018}. A prominent application of \glspl{ODSW} is the generation of training targets for supervised learning of \gls{DSW} estimators. This has been done extensively for audiovisual \gls{ASR}, but is generally application-independent. Hence, a means to obtain \glspl{ODSW} based on the nonlinear \gls{DS} model discussed in the previous section is presented in the following.

\subsection{Maximum likelihood estimation}
\label{subsec:odsw_ml}
The likelihood function introduced in Eq.~\eqref{eqn:lds_joint_likelihood} can be exploited to obtain \glspl{ODSW} if the \gls{DS} in Eqs.~\eqref{eqn:lds_process_model}--\eqref{eqn:lds_observation_model} is fully observed~\cite{Schymura2018}. Therefore, a prior distribution other than the uniform prior has to be imposed on the \glspl{DSW}. If a uniform prior is assumed, the optimization function will be a linear function of the  \glspl{DSW}. This results in a problem already reported in the context of audiovisual \gls{ASR}, where all \glspl{ODSW} are restricted to take boundary values \(\lambda_{m, k}^{\star} \in \{0,\,1\}\), preventing a smooth weighting of the individual modalities~\cite{Abdelaziz2015, Potamianos1998}. Therefore, given a sequence of observed states \(\mathcal{X}_{1 : K} = \{\boldsymbol{x}_{1},\,\ldots,\,\boldsymbol{x}_{K}\}\) and \(M\) observations \(\mathcal{Y}_{m, 1 : K} = \{\boldsymbol{y}_{m, 1},\,\ldots,\,\boldsymbol{y}_{m, K}\}\) with \(m = 1,\,\ldots,\,M\), a modified joint likelihood function for the fully observed model can be expressed as
\begin{align}
p(\mathcal{X}_{1 : K},\,\mathcal{Y}_{1, 1 : K},&\,\ldots,\,\mathcal{Y}_{M, 1 : K},\,\mathcal{L}_{1 : K}) \propto \notag \\ 
&\prod_{k = 1}^{K} \prod_{m = 1}^{M} p(\boldsymbol{y}_{m, k}\,|\,\boldsymbol{x}_{k})^{\lambda_{m, k}} p(\lambda_{m, k}),
\label{eqn:likelihood_function_with_prior}
\end{align}
where \(\mathcal{L}_{k} = \{\boldsymbol{\lambda}_{1},\,\ldots,\,\boldsymbol{\lambda}_{k}\}\) with \(\boldsymbol{\lambda}_{k} = \begin{bmatrix}
\lambda_{1, k} & \cdots & \lambda_{M, k}\end{bmatrix}^{\mathrm{T}}\) is a sequence of \glspl{DSW}, which are i.i.d. and obey the constraint \(\sum_{m = 1}^{M} \lambda_{m, k} = 1~\forall\,k\).

\subsection{Gaussian prior for the special case with two observations}
\label{subsec:odsw_ml_gaussian}

A method to obtain a \gls{ML} estimate of the \glspl{ODSW} with a Gaussian prior was proposed in~\cite{Schymura2018} for the special case of \(M = 2\), which requires a scalar \gls{ODSW} \(\lambda_{k}^{\star} = \lambda_{1, k}^{\star} = 1 - \lambda_{2, k}^{\star}\) per time step. An analytic solution
\begin{equation}
\lambda_{k}^{\star} = \mu_{\lambda} + \sigma_{\lambda}^{2} \log \Big\{\frac{p(\boldsymbol{y}_{1, k}\,|\,\boldsymbol{x}_{k})}{p(\boldsymbol{y}_{2, k}\,|\,\boldsymbol{x}_{k})}\Big\}
\label{eqn:gaussian_odsw}
\end{equation}
for \glspl{LDS} was derived, where \(\mu_{\lambda}\) denotes the mean and \(\sigma_{\lambda}^{2}\) represents the variance of the Gaussian prior. This solution is closely related to the Gaussian \gls{ODSW} estimator for coupled \glspl{HMM} introduced in~\cite{Abdelaziz2015}. The \glspl{ODSW} were clipped to fit into the range \([0,\,1]\). If the mean and variance parameters are appropriately chosen, the resulting distribution could still be approximately assumed as Gaussian within this interval. Nonetheless, a straightforward extension to \glspl{DS} with \(M > 2\) is problematic, as a Gaussian \gls{PDF} is not able to handle the constraint that multiple \glspl{ODSW} have to sum to one. Clipping with subsequent renormalization of the \gls{ODSW} values could be utilized in this case. However, an accessible interpretation of the mean and variance parameters of the Gaussian prior remains unclear, as clipping and renormalization impose a nonlinear transform on the obtained \glspl{ODSW}. The Gaussian prior requires optimization of two hyperparameters on a dedicated validation set, which is usually conducted via a computationally expensive grid search, cf.~\cite{Gergen2016}.

\subsection{Dirichlet prior for an arbitrary number of observations}
\label{subsec:odsw_dirichlet}
To cope with an arbitrary number of observations in a theoretically sound and interpretable probabilistic framework, a symmetric Dirichlet prior
\begin{equation}
p(\lambda_{m, k}) = \frac{\Gamma(\alpha M)}{\Gamma(\alpha)^{M}} \prod_{m = 1}^{M} \lambda_{m, k}^{\alpha - 1},
\label{eqn:dirichlet_prior}
\end{equation}
with concentration parameter \(\alpha > 1\) is utilized in this work. This single hyperparameter still has to be tuned by e.g. a grid search. Inserting Eq.~\eqref{eqn:dirichlet_prior} into Eq.~\eqref{eqn:likelihood_function_with_prior} and taking into account the i.i.d. property of subsequent trajectory points yields
\begin{align}
p(\boldsymbol{x}_{k},\,\boldsymbol{y}_{1, k},\,&\ldots,\,\boldsymbol{y}_{M, k},\,\boldsymbol{\lambda}_{k}) \propto \notag \\ 
&\frac{\Gamma(\alpha M)}{\Gamma(\alpha)^{M}} \prod_{m = 1}^{M} p(\boldsymbol{y}_{m, k}\,|\,\boldsymbol{x}_{k})^{\lambda_{m, k}} \lambda_{m, k}^{\alpha - 1},
\label{eqn:likelihood_function_with_dirichlet_prior}
\end{align}
for the \(k\)-th time step, which can be transformed into the log-domain and hence serve as an objective function for \gls{ML} estimation according to
\begin{align}
J(\boldsymbol{\lambda}_{k}) &= \sum_{m = 1}^{M} \lambda_{m, k} \log\{p(\boldsymbol{y}_{m, k}\,|\,\boldsymbol{x}_{k})\} \notag \\
& + (\alpha - 1) \sum_{m = 1}^{M} \log\{\lambda_{m, k}\} + \mathrm{const.}
\label{eqn:log_likelihood_with_dirichlet_prior}
\end{align}
Obtaining a \gls{ML} estimate of the \glspl{ODSW} therefore requires to solve the optimization problem
\begin{equation}
\boldsymbol{\lambda}_{k}^{\star} = \underset{\boldsymbol{\lambda}_{k}}{\max}~J(\boldsymbol{\lambda}_{k}) \quad \text{subject to}~~\sum_{m = 1}^{M} \lambda_{m, k} = 1~\forall\,k \label{eqn:optimization_problem}
\end{equation}
for each time step. It is shown in Appendix~\ref{sec:app:concave} that the objective function given in Eq.~\eqref{eqn:log_likelihood_with_dirichlet_prior} is strictly concave for \(\alpha > 1\). As the maximization of a concave function is a convex optimization problem, efficient algorithms to solve the problem stated in Eq.~\eqref{eqn:optimization_problem} can be utilized, cf.~\cite[Chap.~3]{Boyd2004}. This also guarantees a unique solution, which corresponds to a global optimum~\cite{Grant2008, Grant2014}.

\section{Dynamic Stream Weight Prediction Models}
\label{sec:learning_framework}
To deploy the proposed state estimation framework in actual application scenarios, a remaining issue has to be solved: \glspl{ODSW} can only be obtained for fully-observed models. Hence, \glspl{DSW} must be estimated from available instantaneous sensor reliability measures. This procedure has already been established for \gls{DSW}-based models in audiovisual \gls{ASR}, where, for instance, the instantaneous estimated acoustic \gls{SNR} was used as such a measure~\cite{Abdelaziz2015}. 

Let \(\hat{\boldsymbol{\lambda}}_{k} = g(\boldsymbol{z}_{k},\,\boldsymbol{\theta})\) denote the general structure of a prediction model with parameters \(\boldsymbol{\theta}\), where the predicted \glspl{DSW} at time step \(k\) are denoted as \(\hat{\boldsymbol{\lambda}}_{k} = \begin{bmatrix} \hat{\lambda}_{1, k} & \cdots & \hat{\lambda}_{M, k}\end{bmatrix}^{\mathrm{T}}\) and \(\boldsymbol{z}_{k}\) is a vector of reliability measures or, more generally, features that describe the instantaneous measurement uncertainty associated with the corresponding sensors. The prediction model can be any nonlinear function, with the constraint that it is differentiable w.r.t. its parameters. Additionally, the individual function outputs must sum to one, cf. Sec.~\ref{subsec:nonlinear_system_model}. Prominent models that match these requirements are e.g. logistic functions or a neural network with softmax output layer.

Supervised training of \gls{DSW} prediction models requires the availability of a training dataset, where ground-truth information about the state \(\mathcal{X}_{\mathrm{train}}\), the corresponding observations \(\mathcal{Y}_{\mathrm{train}}\) and the associated reliability measures \(\mathcal{Z}_{\mathrm{train}}\) are available. Although the Gaussian filtering paradigm utilizes time series data, individual data points used during training can be assumed i.i.d. as \glspl{ODSW} can be estimated independently for each time step, cf. Sec.~\ref{subsec:odsw_dirichlet}.

\begin{figure}[t]
	\centering
	\pgfdeclarelayer{background}
\pgfdeclarelayer{foreground}
\pgfsetlayers{background,main,foreground}

\definecolor{rub_green}{rgb}{0.5804,0.7569,0.1059}%
\definecolor{rub_blue}{rgb}{0,0.2078,0.3765}%
\definecolor{rub_gray}{rgb}{0.8666,0.8666,0.8666}%
\definecolor{rub_orange}{rgb}{0.8941,0.5333,0.2549}

\tikzstyle{block} = [draw, text width=4.5em, text centered, fill=black!02, minimum height=3.5em, drop shadow=black]

\begin{tikzpicture}
	\node (odsw_estimation) [block] {ODSW Estimation};
	\node (parameter_estimation) [block, right of=odsw_estimation, xshift=7em] {Parameter Learning};
	
	\draw [-{Latex[scale=1.0]}] (odsw_estimation.east) -- node[auto] {$\mathcal{L}^{\star}_{\mathrm{train}}$} (parameter_estimation.west);
	
	\node (x_train) [left of=odsw_estimation, node distance = 6em, yshift=1em] {$\mathcal{X}_{\mathrm{train}}$};
	\node (y_train) [left of=odsw_estimation, node distance = 6em, yshift=-1em] {$\mathcal{Y}_{\mathrm{train}}$};
	
	\draw [-{Latex[scale=1.0]}] (x_train) -- node[auto] {} ([yshift=1em]odsw_estimation.west);
	\draw [-{Latex[scale=1.0]}] (y_train) -- node[auto] {} ([yshift=-1em]odsw_estimation.west);
	
	\node (theta) [right of=parameter_estimation, node distance = 5em] {$\boldsymbol{\theta}$};
	
	\draw [-{Latex[scale=1.0]}] (parameter_estimation.east) -- node[auto] {} (theta);
	
	\node (z_train) [above of=parameter_estimation, node distance = 4em] {$\mathcal{Z}_{\mathrm{train}}$};
	
	\draw [-{Latex[scale=1.0]}] (z_train) -- node[auto] {} (parameter_estimation.north);
\end{tikzpicture}%
	\caption{Block diagram of the two-stage training process utilized for obtaining \gls{DSW} prediction model parameters.}
	\label{fig:training_stage}
\end{figure}
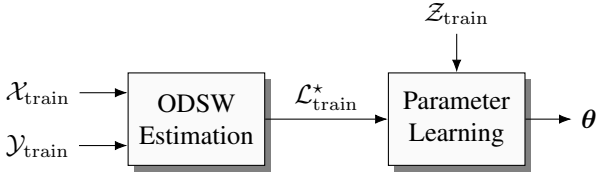

The training phase is a two-stage process, which is illustrated in Fig.~\ref{fig:training_stage}. First, \glspl{ODSW} \(\mathcal{L}^{\star}_{\mathrm{train}}\) are estimated for the available training data using the method described in Sec.~\ref{subsec:odsw_dirichlet}. Subsequently, supervised training of the model parameters is conducted using reliability measures as inputs and \glspl{ODSW} as targets. Due to the constraint that \glspl{DSW} must sum to one and the imposed symmetric Dirichlet prior, the \glspl{ODSW} can be assumed to stem from a categorical distribution. Therefore, appropriate loss functions that can be exploited here are e.g. the \gls{KLD} or the cross-entropy loss~\cite{Wang2018}. This allows to utilize a gradient-based optimizer for the second part of the training phase.

%
\section{Evaluation}
\label{sec:evaluation}

The experimental evaluation in this work focuses on three scenarios: the evaluation of the proposed \gls{ODSW} estimation technique, the comparison of the proposed framework with related state-of-the-art audiovisual speaker tracking methods and an empirical assessment of the computational complexity compared to the standard \gls{EKF}.

\subsection{Audiovisual datasets}
\label{subsec:audiovisual_datasets}

\begin{figure}%
		\begin{minipage}{0.166\textwidth}
%
%
\begin{tikzpicture}

\begin{axis}[%
width=1.1in,
height=0.8in,
at={(0in,0in)},
scale only axis,
axis on top,
xmin=0.5,
xmax=640.5,
xticklabels={,,},
yticklabels={,,},
y dir=reverse,
xshift=-1cm,
ymin=0.5,
ymax=480.5,
ticks=none,
axis background/.style={fill=white},
legend style={legend cell align=left, align=left, draw=white!15!black}
]
\addplot [forget plot] graphics [xmin=0.5, xmax=640.5, ymin=0.5, ymax=480.5] {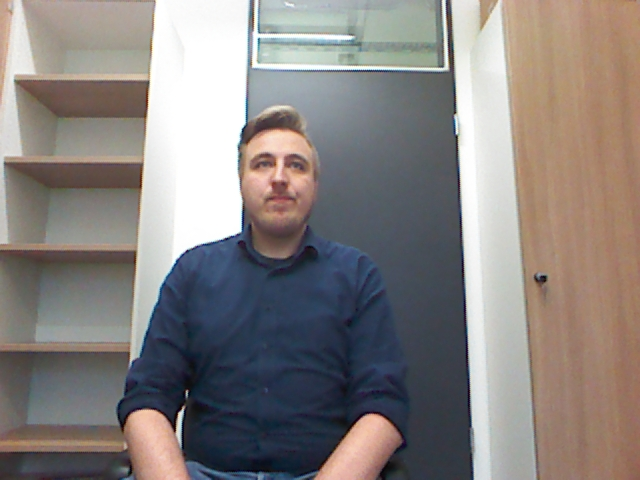};
 \end{axis}
\end{tikzpicture}%
		\end{minipage}%
		\begin{minipage}{0.166\textwidth}
%
%
\begin{tikzpicture}

\begin{axis}[%
width=1.1in,
height=0.8in,
at={(0in,0in)},
scale only axis,
axis on top,
xmin=0.5,
xmax=320.5,
xticklabels={,,},
yticklabels={,,},
y dir=reverse,
xshift=-1cm,
ymin=0.5,
ymax=240.5,
ticks=none,
axis background/.style={fill=white},
legend style={legend cell align=left, align=left, draw=white!15!black}
]
\addplot [forget plot] graphics [xmin=0.5, xmax=320.5, ymin=0.5, ymax=240.5] {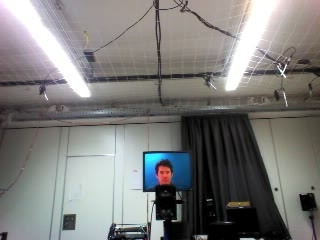};
\end{axis}
\end{tikzpicture}%
		\end{minipage}%
		\begin{minipage}{0.166\textwidth}
%
%
\begin{tikzpicture}

\begin{axis}[%
width=1.1in,
height=0.8in,
at={(0in,0in)},
scale only axis,
axis on top,
xmin=0.5,
xmax=640.5,
xticklabels={,,},
yticklabels={,,},
y dir=reverse,
xshift=-1cm,
ymin=0.5,
ymax=480.5,
ticks=none,
axis background/.style={fill=white},
legend style={legend cell align=left, align=left, draw=white!15!black}
]
\addplot [forget plot] graphics [xmin=0.5, xmax=640.5, ymin=0.5, ymax=480.5] {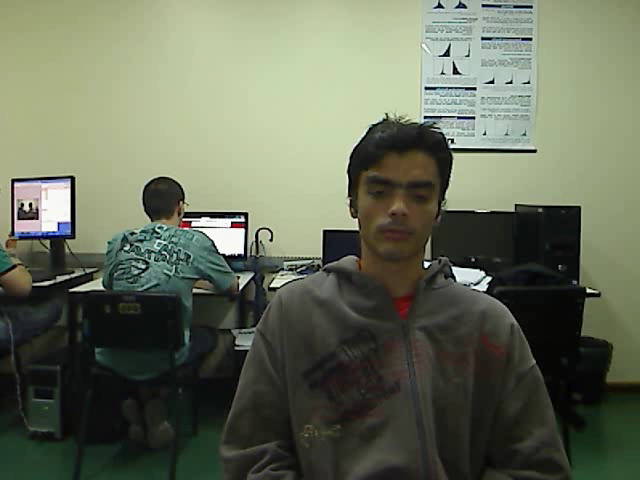};
\end{axis}
\end{tikzpicture}%
		\end{minipage}\vspace{0.25cm}
	
		\begin{minipage}{0.166\textwidth}
%
%
\begin{tikzpicture}

\begin{axis}[%
width=1.1in,
height=0.8in,
at={(0in,0in)},
scale only axis,
axis on top,
xmin=0.5,
xmax=640.5,
xticklabels={,,},
yticklabels={,,},
y dir=reverse,
xshift=-1cm,
ymin=0.5,
ymax=480.5,
ticks=none,
axis background/.style={fill=white},
legend style={legend cell align=left, align=left, draw=white!15!black}
]
\addplot [forget plot] graphics [xmin=0.5, xmax=640.5, ymin=0.5, ymax=480.5] {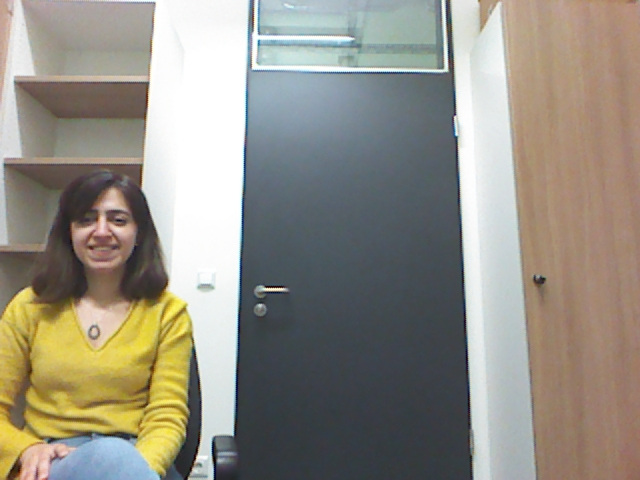};
\end{axis}
\end{tikzpicture}%
		\end{minipage}%
		\begin{minipage}{0.166\textwidth}
%
%
\begin{tikzpicture}

\begin{axis}[%
width=1.1in,
height=0.8in,
at={(0in,0in)},
scale only axis,
axis on top,
xmin=0.5,
xmax=320.5,
xticklabels={,,},
yticklabels={,,},
y dir=reverse,
xshift=-1cm,
ymin=0.5,
ymax=240.5,
ticks=none,
axis background/.style={fill=white},
legend style={legend cell align=left, align=left, draw=white!15!black}
]
\addplot [forget plot] graphics [xmin=0.5, xmax=320.5, ymin=0.5, ymax=240.5] {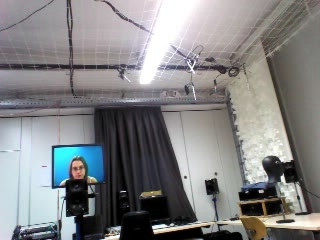};
\end{axis}
\end{tikzpicture}%
		\end{minipage}%
		\begin{minipage}{0.166\textwidth}
%
%
\begin{tikzpicture}

\begin{axis}[%
width=1.1in,
height=0.8in,
at={(0in,0in)},
scale only axis,
axis on top,
xmin=0.5,
xmax=960.5,
xticklabels={,,},
yticklabels={,,},
y dir=reverse,
xshift=-1cm,
ymin=0.5,
ymax=720.5,
ticks=none,
axis background/.style={fill=white},
legend style={legend cell align=left, align=left, draw=white!15!black}
]
\addplot [forget plot] graphics [xmin=0.5, xmax=960.5, ymin=0.5, ymax=720.5] {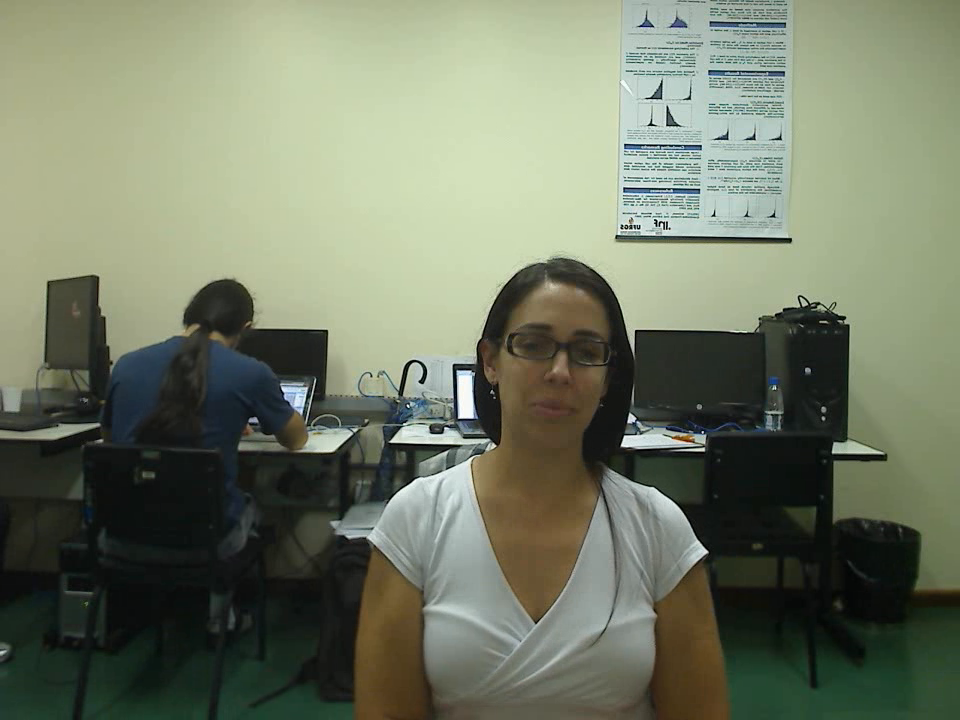};
\end{axis}
\end{tikzpicture}%
		\end{minipage}%
	\caption{Exemplary still images from the \gls{KAVLoC} (left column), \gls{NAVLoC} (center column) and \gls{MVAD} (right column) datasets.}
	\label{fig:database_frames}
\end{figure}

Three audiovisual datasets are used to conduct the experiments in this study. These datasets comprise a variety of recording conditions and were acquired using different acoustic and visual sensors. The evaluation data was purposefully selected to reflect a wide range of application scenarios with different dynamics and sensor disturbances. To give a first impression of the dataset variability, some exemplary still images from the video files of the evaluation corpora are depicted in Fig.~\ref{fig:database_frames}.

The first audiovisual corpus was specifically recorded for this study and will be referred to as the \gls{KAVLoC}. This dataset contains audiovisual recordings of seven (four male and three female) speakers, acquired in an office room with an average reverberation time of approximately \(200\,\mathrm{ms}\). A Microsoft Kinect\texttrademark~sensor was positioned on a table at a height of \(0.9\,\mathrm{m}\). The participants were sitting on a chair facing the sensor at a distance of approximately \(1.5\,\mathrm{m}\). Besides being advised to stay seated, they were allowed to move freely during the recordings. Throughout each recording session, the speakers were asked to read out sentences randomly selected from the CSTR VCTK corpus~\cite{Veaux2017}, which is composed of over 400 sentences taken from English newspapers. Ten audiovisual sequences of \(30\,\mathrm{s}\) duration at two different positions were recorded for each speaker. Acoustic signals were acquired at a frame rate of \(16\,\mathrm{kHz}\) using the four-channel microphone array of the Kinect\texttrademark~sensor. The corresponding video sequences were recorded with a resolution of \(640 \times 480\) pixels at a rate of \(15\)~\gls{FPS}. To obtain the ground-truth speaker locations, the position of the speakers faces were manually annotated in the recorded video signals. The total duration of all acquired audiovisual sequences is \(35\,\mathrm{min}\).

The \gls{NAVLoC} is used as a second dataset. It was already used in a previous work on \glspl{DSW} for \glspl{LDS}~\cite{Schymura2018}. The audiovisual recordings in this dataset were obtained using the humanoid robot NAO in a laboratory environment with an average reverberation time of approximately \(450\,\mathrm{ms}\). A computer screen and a loudspeaker were positioned at a distance of \(2\,\mathrm{m}\) from the robot. The screen was placed at the same height as the robot's head. Audiovisual sequences from two male and two female speakers were selected randomly from the GRID corpus~\cite{Cooke06} and played back over the screen and the loudspeaker. For each speaker, 100 sequences with a duration of \(2.5\,\mathrm{s}\) each were recorded using the four-channel microphone array and the upper camera of the NAO robot. For one half of these sequences, the robot was directly facing the screen, whereas for the other half, the robot's head was turned \(21^{\circ}\) to the right to enforce a different relative azimuth to the speaker. A sampling rate of \(48\,\mathrm{kHz}\) was used for the acoustic recordings. The video signals were acquired with a resolution of \(320 \times 240\) pixels at \(10\)~\gls{FPS}. As the ground-truth azimuth is directly related to the heading direction of the robot's head, a manual annotation of the collected audiovisual data was not required. The total duration of the recorded sequences in the \gls{NAVLoC} dataset is approximately \(17\,\mathrm{min}\). As the captured microphone signals are corrupted by fan noise of the NAO robot, this dataset is especially challenging regarding the acoustic localization performance.

The dataset for \gls{MVAD} introduced in~\cite{Minotto2015} serves as the third evaluation corpus in this study. It provides audiovisual sequences of single and multiple speakers in an office environment. The recordings were acquired using a Kinect\texttrademark~sensor for capturing the video signals, whereas the audio was captured with an eight-channel linear microphone array. The audio sampling rate is \(44.1\,\mathrm{kHz}\) and the video resolution is \(640 \times 480\) pixels at \(10\) \gls{FPS}. The duration of the individual recordings ranges from \(40\,\mathrm{s}\) to \(60\,\mathrm{s}\), with silent periods of \(4\,\mathrm{s}\) to \(8\,\mathrm{s}\) in between speech segments. Throughout the recordings, the speakers always face the camera and their position changes only slightly. Out of the 31 audiovisual sequences provided in total, six recordings, where only a single speaker was present, were utilized for the experimental evaluation in this study.

\subsection{Evaluation metrics and significance tests}
\label{subsec:evaluation_metrics}
To assess the speaker localization and tracking performance, the circular \gls{RMSE} 
\begin{equation}
\mathrm{cRMSE} = \sqrt{\frac{1}{K - k_{0} + 1} \sum_{k = k_{0}}^{K} \underset{l \in [-\infty,\,\infty]}{\min} \Big(\hat{\phi}_{k} - \phi_{k} + 2\pi l\Big)^{2}}
\label{eqn:circular_rmse}
\end{equation}
was employed as an evaluation metric~\cite{Traa2013}, where \(\hat{\phi}_{k}\) is the estimated azimuth at time-step \(k\), \(\phi_{k}\) is the corresponding ground-truth azimuth angle, \(K\) is the total number of time-steps in one test sequence and \(k_{0}\) corresponds to the number of frames in the grace period. This metric was calculated individually for each audiovisual test sequence. A grace period with \(10\,\%\) of the total sequence length was excluded at the beginning of each sequence to allow the Bayesian filtering frameworks to converge. A one-way \gls{ANOVA} with Bonferroni correction~\cite{Dunn1961} was used to assess statistical significance in all conducted experiments.

\subsection{Experimental setup}
\label{subsec:experimental_setup}
Throughout all experiments, acoustic signals were processed using an acoustic front-end, which includes an initial \gls{VAD}, followed by an instantaneous estimation of the \gls{SNR} and the actual speaker localization. All processing steps were conducted frame-wise at time intervals matching the corresponding video frame rate.

The \gls{VAD}~\cite{Sohn1999} operates on the first channel of the available microphone signals to distinguish between speech and silence frames. Acoustic localization was performed during speech segments only and skipped otherwise. To obtain the instantaneous \gls{SNR}, the unbiased \gls{MMSE} estimator proposed in~\cite{Gerkmann2012} was used to estimate the noise power at each time-frequency point. The noise power estimate was also used to enhance the noisy speech signals via Wiener filtering. The gross \gls{SNR} \(\xi_{k} \in \mathbb{R}\) averaged over all channels and frequencies was computed as a reliability measure, corresponding to the acoustic sensor uncertainty. Subsequent acoustic localization was performed on the enhanced speech segments using the \gls{SRP-PHAT} algorithm~\cite{Cobos2011}. Visual locations of the speaker's face were extracted from the recorded video using the Viola-Jones algorithm~\cite{Viola2001} and converted to azimuth angles based on the calibrated camera images. A visual reliability measure, indicating a potential rotation of the speaker's head, was derived from the detected face region by horizontally mirroring the image and computing the correlation coefficient \(\rho_{k} \in [-1,\,1]\) between the original and the mirrored image~\cite{Schoenherr2016}. Therefore, the individual vectors of reliability measures used for training the \gls{DSW} prediction model can be expressed as \(\boldsymbol{z}_{k} = \begin{bmatrix} \xi_{k} & \rho_{k}\end{bmatrix}^{\mathrm{T}}\). 
%
%

A constant velocity model~\cite{Vermaak2001} was used to model the system dynamics
\begin{equation}
\boldsymbol{x}_{k} = \begin{bmatrix}
1 & T \\
0 & 1
\end{bmatrix}\,\boldsymbol{x}_{k - 1} + \boldsymbol{v}_{k}
\label{eqn:experiments_process_model}
\end{equation}
with
\begin{equation*}
\boldsymbol{v}_{k} \sim \mathcal{N} \Big(\boldsymbol{0},\,\sigma_{v}^{2} \begin{bmatrix}
\frac{1}{3}T^{3} & \frac{1}{2}T^{2} \\
\frac{1}{2}T^{2} & T
\end{bmatrix}\Big),
\end{equation*}
where \(T\) denotes the time between two consecutive discrete time-steps, \(\sigma_{v}^{2} = 0.3\) is a constant factor and the system state \(\boldsymbol{x}_{k} = \begin{bmatrix} \phi_{k} & \dot{\phi}_{k}\end{bmatrix}^{\mathrm{T}}\) is encoded as the azimuthal speaker position \(\phi_{k}\) and velocity \(\dot{\phi}_{k}\), respectively. As both acoustic and visual sensors directly observe angular values, a \gls{RVM}~\cite{Nies2008} represents the circular nature of observed azimuth angles as
\begin{equation}
\boldsymbol{y}_{m, k} = \begin{bmatrix}
\cos(\phi_{k}) \\
\sin(\phi_{k})
\end{bmatrix} + \boldsymbol{w}_{m, k}, ~\boldsymbol{w}_{m, k} \sim \mathcal{N}(\boldsymbol{0},\,\sigma_{w, m}^{2}\boldsymbol{I}),
\label{eqn:experiments_observation_model}
\end{equation}
where \(\sigma_{w, m}^{2} = 0.01\) denotes the observation noise variance of the \(m\)-th sensor and \(m = \{1,\,2\}\). It should be noted that the system dynamics are based on a linear model. Hence, the standard \gls{KF} prediction step can be exploited here. However, the nonlinear observation models must be handled using the corresponding Jacobians to perform \gls{EKF}-based updates.

The logistic function is utilized as a \gls{DSW} prediction model
\begin{equation}
\lambda_{1, k} = \frac{1}{1 + \exp(\boldsymbol{z}_{k}^{\mathrm{T}} \boldsymbol{w} + b)},
\label{eqn:logistic_function}
\end{equation}
where \(\boldsymbol{w} \in \mathbb{R}^{2}\) is the weight vector and \(b \in \mathbb{R}\) is a bias term. This allows to express the \gls{DSW} prediction model parameters as \(\boldsymbol{\theta} \in \{\boldsymbol{w},\,b\}\). As the number of independent observations is fixed as \(M = 2\) throughout all experiments, it is sufficient to only predict the first (acoustic) \gls{DSW} using Eq.~\eqref{eqn:logistic_function}, as the second (visual) \gls{DSW} is defined as \(\lambda_{2, k} = 1 - \lambda_{1, k}\). The model is trained by minimizing the cross-entropy loss using standard \gls{SGD}.

\section{Results and Discussion}
\label{sec:results_and_discussion}
This section describes the results obtained for the three evaluation scenarios investigated in this study. It should be noted that all conducted experiments focus on single-speaker scenarios only. This restriction was chosen on purpose, as it allows to exclusively focus on the localization and tracking performance, without taking into account additional external factors like data association ambiguities, estimating the number of speakers and track-to-track fusion. Multi-speaker tracking is an important issue that must be taken into account for many potential applications. However, as the proposed framework is based on the conventional \gls{EKF} paradigm, it can be easily extended using existing probabilistic data association techniques, cf.~\cite{Chakrabarty2014}. This is outside the scope of this study and will be investigated in future work.

\begin{table*}
	\caption{Circular root mean squared errors in degrees with corresponding standard deviations, obtained by the proposed Bayesian filtering framework using oracle dynamic stream weights and the extended Kalman filter baseline methods. Stars (\(^{\star}\)) indicate a statistically significant improvement of the \gls{ODSW}-\glspl{EKF} over the \gls{EKF} baseline with \(p < 0.05\)}
	\centering
	\renewcommand{\arraystretch}{1.5}
	\footnotesize
	\begin{tabular}{@{}l@{}c@{}r@{}c@{}rrr@{}c@{}rrr@{}}
		\toprule
		& \phantom{aa} & \multicolumn{1}{c}{Undistorted} & \phantom{aa} & \multicolumn{3}{c}{Signal-to-noise ratio} & \phantom{aa} & \multicolumn{3}{c}{Image rotation} \\
		\cmidrule{3-3} \cmidrule{5-7} \cmidrule{9-11}
		& \phantom{aa} & & \phantom{aa} &\multicolumn{1}{c}{\(0\,\mathrm{dB}\)} & \multicolumn{1}{c}{\(15\,\mathrm{dB}\)} & \multicolumn{1}{c}{\(30\,\mathrm{dB}\)} & \phantom{aa} & \multicolumn{1}{c}{\(10^{\circ}\)} & \multicolumn{1}{c}{\(15^{\circ}\)} & \multicolumn{1}{c}{\(20^{\circ}\)} \\
		\midrule
		\textbf{KAVLoC} (\(N = 70\)) \\
		EKF (Audio) & \phantom{a} & \(\phantom{1}5.34 \pm 1.80\)\phantom{\(^{\star}\)} & \phantom{a} & \(11.53 \pm 5.02\)\phantom{\(^{\star}\)} & \(\phantom{1}5.74 \pm 1.90\)\phantom{\(^{\star}\)} & \(\phantom{1}5.30 \pm 1.79\)\phantom{\(^{\star}\)} & \phantom{a} & \(\phantom{1}5.34 \pm 1.80\)\phantom{\(^{\star}\)} & \(\phantom{1}5.34 \pm 1.80\)\phantom{\(^{\star}\)} & \(\phantom{1}5.34 \pm 1.80\)\phantom{\(^{\star}\)} \\ 
		EKF (Video) & \phantom{a} & \(\phantom{1}5.19 \pm 4.12\)\phantom{\(^{\star}\)} & \phantom{a} & \(\phantom{1}5.19 \pm 4.12\)\phantom{\(^{\star}\)} & \(\phantom{1}5.19 \pm 4.12\)\phantom{\(^{\star}\)} & \(\phantom{1}5.19 \pm 4.12\)\phantom{\(^{\star}\)} & \phantom{a} & \(\phantom{1}6.09 \pm 1.41\)\phantom{\(^{\star}\)} & \(\phantom{1}7.36 \pm 1.84\)\phantom{\(^{\star}\)} & \(9.51 \pm 3.82\)\phantom{\(^{\star}\)} \\ 
		EKF (Audiovisual) & \phantom{a} & \(\phantom{1}4.77 \pm 2.89\)\phantom{\(^{\star}\)} & \phantom{a} & \(\phantom{1}6.69 \pm 3.30\)\phantom{\(^{\star}\)} & \(\phantom{1}4.99 \pm 2.86\)\phantom{\(^{\star}\)} & \(\phantom{1}4.78 \pm 2.91\)\phantom{\(^{\star}\)} & \phantom{a} & \(\phantom{1}5.06 \pm 1.62\)\phantom{\(^{\star}\)} & \(\phantom{1}5.58 \pm 1.49\)\phantom{\(^{\star}\)} & \(\phantom{1}6.39 \pm 1.68\)\phantom{\(^{\star}\)} \\ 
		ODSW-EKF (Gaussian) & \phantom{a} & \(\phantom{1}4.25 \pm 1.67\)\phantom{\(^{\star}\)} & \phantom{a} & \(\phantom{1}4.37 \pm 1.74\)\(^{\star}\) & \(\phantom{1}4.27 \pm 1.67\)\phantom{\(^{\star}\)} & \(\phantom{1}4.25 \pm 1.74\)\phantom{\(^{\star}\)} & \phantom{a} & \(\phantom{1}4.74 \pm 1.33\)\phantom{\(^{\star}\)} & \(\phantom{1}5.00 \pm 1.34\)\phantom{\(^{\star}\)} & \(\phantom{1}5.35 \pm 1.54\)\(^{\star}\) \\ 
		ODSW-EKF (Dirichlet) & \phantom{a} & \(\phantom{1}4.15 \pm 1.38\)\phantom{\(^{\star}\)} & \phantom{a} & \(\phantom{1}4.28 \pm 1.47\)\(^{\star}\) & \(\phantom{1}4.18 \pm 1.39\)\phantom{\(^{\star}\)} & \(\phantom{1}4.15 \pm 1.41\)\phantom{\(^{\star}\)} & \phantom{a} & \(\phantom{1}4.81 \pm 1.34\)\phantom{\(^{\star}\)} & \(\phantom{1}5.07 \pm 1.38\)\phantom{\(^{\star}\)} & \(\phantom{1}5.40 \pm 1.65\)\(^{\star}\) \\ 
		\textbf{NAVLoC} (\(N = 400\)) \\
		EKF (Audio) & \phantom{a} & \(10.86 \pm 3.99\)\phantom{\(^{\star}\)} & \phantom{a} & \(10.65 \pm 3.36\)\phantom{\(^{\star}\)} & \(10.79 \pm 3.86\)\phantom{\(^{\star}\)} & \(10.85 \pm 3.96\)\phantom{\(^{\star}\)} & \phantom{a} & \(10.86 \pm 3.99\)\phantom{\(^{\star}\)} & \(10.86 \pm 3.99\)\phantom{\(^{\star}\)} & \(10.86 \pm 3.99\)\phantom{\(^{\star}\)} \\ 
		EKF (Video) & \phantom{a} & \(\phantom{1}8.82 \pm 0.70\)\phantom{\(^{\star}\)} & \phantom{a} & \(\phantom{1}8.82 \pm 0.70\)\phantom{\(^{\star}\)} & \(\phantom{1}8.82 \pm 0.70\)\phantom{\(^{\star}\)} & \(\phantom{1}8.82 \pm 0.70\)\phantom{\(^{\star}\)} & \phantom{a} & \(\phantom{1}9.12 \pm 1.53\)\phantom{\(^{\star}\)} & \(9.86 \pm 1.83\)\phantom{\(^{\star}\)} & \(9.81 \pm 1.30\)\phantom{\(^{\star}\)} \\ 
		EKF (Audiovisual) & \phantom{a} & \(9.54 \pm 2.82\)\phantom{\(^{\star}\)} & \phantom{a} & \(\phantom{1}9.40 \pm 2.44\)\phantom{\(^{\star}\)} & \(\phantom{1}9.49 \pm 2.74\)\phantom{\(^{\star}\)} & \(9.53 \pm 2.81\)\phantom{\(^{\star}\)} & \phantom{a} & \(9.94 \pm 3.40\)\phantom{\(^{\star}\)} & \(10.50 \pm 3.90\)\phantom{\(^{\star}\)} & \(10.65 \pm 3.98\)\phantom{\(^{\star}\)} \\ 
		ODSW-EKF (Gaussian) & \phantom{a} & \(\phantom{1}8.83 \pm 0.81\)\(^{\star}\) & \phantom{a} & \(\phantom{1}8.82 \pm 0.76\)\(^{\star}\) & \(\phantom{1}8.83 \pm 0.81\)\(^{\star}\) & \(\phantom{1}8.83 \pm 0.82\)\(^{\star}\) & \phantom{a} & \(\phantom{1}9.00 \pm 1.79\)\(^{\star}\) & \(9.72 \pm 2.82\)\(^{\star}\) & \(10.16 \pm 3.31\)\phantom{\(^{\star}\)} \\ 
		ODSW-EKF (Dirichlet) & \phantom{a} & \(\phantom{1}8.83 \pm 0.81\)\(^{\star}\) & \phantom{a} & \(\phantom{1}8.81 \pm 0.76\)\(^{\star}\) & \(\phantom{1}8.82 \pm 0.81\)\(^{\star}\) & \(\phantom{1}8.83 \pm 0.82\)\(^{\star}\) & \phantom{a} & \(\phantom{1}8.99 \pm 1.79\)\(^{\star}\) & \(9.72 \pm 2.82\)\(^{\star}\) & \(10.16 \pm 3.31\)\phantom{\(^{\star}\)} \\
		\textbf{MVAD} (\(N = 6\)) \\
		EKF (Audio) & \phantom{a} & \(\phantom{1}5.37 \pm 2.53\)\phantom{\(^{\star}\)} & \phantom{a} & \(10.33 \pm 5.69\)\phantom{\(^{\star}\)} & \(10.73 \pm 6.65\)\phantom{\(^{\star}\)} & \(\phantom{1}4.81 \pm 1.84\)\phantom{\(^{\star}\)} & \phantom{a} & \(\phantom{1}5.37 \pm 2.53\)\phantom{\(^{\star}\)} & \(\phantom{1}5.37 \pm 2.53\)\phantom{\(^{\star}\)} & \(\phantom{1}5.37 \pm 2.53\)\phantom{\(^{\star}\)} \\ 
		EKF (Video) & \phantom{a} & \(\phantom{1}1.81 \pm 1.69\)\phantom{\(^{\star}\)} & \phantom{a} & \(\phantom{1}1.81 \pm 1.69\)\phantom{\(^{\star}\)} & \(\phantom{1}1.81 \pm 1.69\)\phantom{\(^{\star}\)} & \(\phantom{1}1.81 \pm 1.69\)\phantom{\(^{\star}\)} & \phantom{a} & \(\phantom{1}4.33 \pm 2.41\)\phantom{\(^{\star}\)} & \(\phantom{1}3.89 \pm 1.37\)\phantom{\(^{\star}\)} & \(\phantom{1}4.89 \pm 1.81\)\phantom{\(^{\star}\)} \\ 
		EKF (Audiovisual) & \phantom{a} & \(\phantom{1}2.32 \pm 1.50\)\phantom{\(^{\star}\)} & \phantom{a} & \(\phantom{1}2.98 \pm 1.53\)\phantom{\(^{\star}\)} & \(\phantom{1}2.34 \pm 1.59\)\phantom{\(^{\star}\)} & \(\phantom{1}2.33 \pm 1.55\)\phantom{\(^{\star}\)} & \phantom{a} & \(\phantom{1}4.31 \pm 2.06\)\phantom{\(^{\star}\)} & \(\phantom{1}4.16 \pm 0.74\)\phantom{\(^{\star}\)} & \(\phantom{1}5.00 \pm 1.41\)\phantom{\(^{\star}\)} \\ 
		ODSW-EKF (Gaussian) & \phantom{a} & \(\phantom{1}1.71 \pm 1.65\)\phantom{\(^{\star}\)} & \phantom{a} & \(\phantom{1}1.76 \pm 1.64\)\phantom{\(^{\star}\)} & \(\phantom{1}1.72 \pm 1.65\)\phantom{\(^{\star}\)} & \(\phantom{1}1.71 \pm 1.65\)\phantom{\(^{\star}\)} & \phantom{a} & \(\phantom{1}3.99 \pm 2.58\)\phantom{\(^{\star}\)} & \(\phantom{1}3.87 \pm 1.36\)\phantom{\(^{\star}\)} & \(\phantom{1}5.04 \pm 2.12\)\phantom{\(^{\star}\)} \\ 
		ODSW-EKF (Dirichlet) & \phantom{a} & \(\phantom{1}1.71 \pm 1.65\)\phantom{\(^{\star}\)} & \phantom{a} & \(\phantom{1}1.77 \pm 1.65\)\phantom{\(^{\star}\)} & \(\phantom{1}1.72 \pm 1.65\)\phantom{\(^{\star}\)} & \(\phantom{1}1.71 \pm 1.66\)\phantom{\(^{\star}\)} & \phantom{a} & \(\phantom{1}4.00 \pm 2.58\)\phantom{\(^{\star}\)} & \(\phantom{1}3.93 \pm 1.42\)\phantom{\(^{\star}\)} & \(\phantom{1}5.04 \pm 2.12\)\phantom{\(^{\star}\)} \\ 
		\bottomrule
	\end{tabular}
	\label{tab:oracle_results}
\end{table*}

\subsection{Oracle dynamic stream weight performance}
\label{subsec:results_oracle}
The first evaluation scenario focuses on the \gls{ODSW} estimation technique proposed in this study. To analyze the tracking performance under different sensor reliability conditions, the audiovisual signals from all three datasets are augmented with systematic disturbances. Following the approach introduced in~\cite{Schoenherr2016}, the acoustic signals are perturbed with diffuse white noise at different \gls{SNR} levels (\(0\,\mathrm{dB}\), \(15\,\mathrm{dB}\) and \(30\,\mathrm{dB}\)), calculated over each sequence. Image rotations of \(10^{\circ}\), \(15^{\circ}\) and \(20^{\circ}\) are used to simulate disturbances of the visual modality. The average circular \gls{RMSE} is evaluated for each audiovisual sequence in each condition. The standard \gls{EKF} with audiovisual observations serves as the baseline. Results for the single-modality \gls{EKF} with either audio-only or video-only observations are also analyzed for comparison. The proposed \gls{ODSW} estimation framework is assessed using both the Gaussian prior as proposed in~\cite{Schymura2018}, as well as the Dirichlet prior from this work. All experiments were performed following a leave-one-out cross-validation scheme. Tab.~\ref{tab:oracle_results} summarizes the results achieved in this evaluation scenario.

The results obtained for the single-modality \gls{EKF} baselines in the \gls{KAVLoC} dataset indicates that both acoustic and visual localization achieve similar performance for this corpus. Audiovisual fusion using the standard \gls{EKF} slightly improves localization accuracy over the individual performances, which suggests that fusing both modalities proves to be beneficial using this dataset. The \gls{ODSW}-\gls{EKF} further improves performance compared to the audiovisual \gls{EKF}. However, this improvement is not statistically significant in the undistorted case and there is only a slight difference between \glspl{ODSW} obtained with a Gaussian prior and the proposed Dirichlet-prior based \gls{ODSW}-\gls{EKF}. Statistically significant improvements were obtained in situations with large disturbances, e.g. \(0\,\mathrm{dB}\) \gls{SNR} and \(20^{\circ}\) image rotation. This observation supports the hypothesis that, without proper adaptation, the standard \gls{EKF} is unable to handle large sensor disturbances effectively. This effect can be observed in all evaluated conditions for this dataset: the performance improvement of the \gls{ODSW}-\gls{EKF} over the \gls{EKF} baseline increases with increasing difference between the single-modality cases.

The particular challenge of the \gls{NAVLoC} dataset is that both acoustic and visual sensors are already affected by significant disturbances, even in the undistorted case. This is primarily caused by fan noise and reverberation for the audio and low image resolution and bright lighting conditions for the video signal. Hence, the systematic disturbances added to the raw audiovisual signals only have little effect, which is reflected by the results obtained for the single-modality \glspl{EKF}. Both \gls{ODSW}-\glspl{EKF} yield statistically significant improvements over the audiovisual \gls{EKF} baseline in all cases except for an image rotation of \(20^{\circ}\). However, the audiovisual \gls{EKF} is also outperformed by the video-only \gls{EKF} in some conditions, which even achieves localization performance similar to the \gls{ODSW}-\glspl{EKF}. This leads to the conclusion that in cases where all available sensors suffer from large disturbances, the standard \gls{EKF} again fails to perform efficient sensor fusion without adaptation. Additionally, the \gls{ODSW}-\gls{EKF} is able to cope with this situation, but is limited by the performance of the best-performing modality.

An improved performance of the \gls{ODSW}-\gls{EKF} in terms of the mean azimuth error can also be observed for the \gls{MVAD} corpus, but due to the small sample size, it is not possible to show statistical significance. A comparison of the single-modality results for this dataset indicates that the visual modality has a largely improved reliability over the acoustic sensors. This again leads to a slightly degraded performance of the audiovisual \gls{EKF}. A reduced localization error compared to the \gls{EKF} baseline and the single-modality \glspl{EKF} is achieved by both \gls{ODSW}-\glspl{EKF} in conditions with acoustic disturbance. A disturbance of the visual modality leads to similar performance for all evaluated methods.

For all evaluated datasets, only marginal performance differences between the Gaussian prior-based \gls{ODSW}-\gls{EKF} and the proposed Dirichlet prior are present. This indicates that both methods are capable of producing reliable \gls{ODSW} estimates. However, as discussed in Sec.~\ref{sec:odsw_dirichlet}, the proposed Dirichlet prior has a plausible probabilistic interpretation and only requires the tuning of a single hyperparameter. Furthermore, empirical observations during the experiments indicated that the method is insensitive to the choice of the concentration parameter to a certain degree. However, this has not been systematically evaluated.


\begin{table}
	\caption{Circular root mean squared errors in degrees with standard deviations obtained using different audiovisual speaker tracking algorithms. Values in a column suffixed with different superscript letters are significantly different from each other at \(p < 0.05\).}
	\centering
	\renewcommand{\arraystretch}{1.5}
	\footnotesize
	\begin{tabular}{@{}l@{}ccc@{}}
		\toprule
		& KAVLoC & NAVLoC & MVAD \\
		\midrule
		EKF & \(\phantom{1}6.16 \pm 1.73^{\mathrm{a}}\) & \(10.08 \pm 3.37^{\mathrm{a}}\) & \(\phantom{1}4.05 \pm 0.61^{\mathrm{a}}\) \\ 
		Gehrig et al.~\cite{Gehrig2005} & \(\phantom{1}6.42 \pm 1.56^{\mathrm{a}}\) & \(10.37 \pm 1.84^{\mathrm{a}}\) & \(\phantom{1}4.59 \pm 0.63^{\mathrm{a}}\) \\ 
		Gerlach et al.~\cite{Gerlach2012} & \(\phantom{1}6.22 \pm 4.46^{\mathrm{a}}\) & \(15.20 \pm 4.68^{\mathrm{b}}\) & \(\phantom{1}2.85 \pm 0.53^{\mathrm{b}}\) \\ 
		Qian et al.~\cite{Qian2017} & \(\phantom{1}6.21 \pm 2.86^{\mathrm{a}}\) & \(10.17 \pm 7.22^{\mathrm{a}}\) & \(\phantom{1}3.93 \pm 0.37^{\mathrm{b}}\) \\ 
		ODSW-EKF (Dirichlet) & \(\phantom{1}5.09 \pm 1.27^{\mathrm{b}}\) & \(\phantom{1}9.32 \pm 1.99^{\mathrm{c}}\) & \(\phantom{1}3.64 \pm 0.91^{\mathrm{b}}\) \\ 
		DSW-EKF & \(\phantom{1}6.12 \pm 1.58^{\mathrm{a}}\) & \(\phantom{1}9.76 \pm 1.99^{\mathrm{a}}\) & \(\phantom{1}3.64 \pm 0.64^{\mathrm{b}}\) \\ 
		\bottomrule
	\end{tabular}
	\label{tab:comparison_results}
\end{table}

\subsection{Audiovisual tracking performance analysis}
\label{subsec:results_comparison}
A comparison of the Bayesian filtering framework proposed in this study with state-of-the-art audiovisual speaker tracking methods is the primary focus of the second evaluation scenario. Four different frameworks were selected as baseline methods: the standard \gls{EKF} with audiovisual observations, the audiovisual fusion technique based on an iterated \gls{EKF} as proposed by Gehring et al.~\cite{Gehrig2005}, the \gls{PF}-based approach with adaptive particle weighting introduced by Gerlach et al.~\cite{Gerlach2012} and the recently proposed framework by Qian et al.~\cite{Qian2017}, which explicitly incorporates sensor reliability measures into the weighting stage of the \gls{PF}. These methods are compared with the \gls{ODSW}-\gls{EKF} with Dirichlet prior and a \gls{DSW}-\gls{EKF} with corresponding prediction model based on the logistic function, as introduced in Sec.~\ref{sec:learning_framework}. The model utilizes the acoustic and visual reliability measures described in Sec.~\ref{subsec:experimental_setup}. A leave-one-out cross-validation procedure identical to the first evaluation scenario is utilized here. The audiovisual sequences of one speaker served as a test set and the sequences from all other speakers were used for training and validation. This procedure was repeated for all speakers in each dataset. More sophisticated realizations of the \gls{DSW} prediction model, e.g. a neural network, can also be exploited for this task. However, initial experiments performed throughout the course of this study have shown that models with increased complexity do not yield any significant benefit over the logistic function utilized here, given the limited set of provided reliability measures. A thorough analysis of specific reliability measures or even the end-to-end training of \gls{DSW} prediction models are beyond the scope of this work. The results for all evaluated methods are shown in Tab.~\ref{tab:comparison_results}, where the achieved circular \glspl{RMSE} are averaged over all systematic disturbance conditions and cross-validation folds.

For the \gls{KAVLoC} dataset, the proposed \gls{ODSW}-\gls{EKF} with Dirichlet prior outperforms all baseline methods with a statistically significant performance benefit. As this result is achieved using a fully observed model, it can be considered an upper bound on performance using this method, which cannot be challenged by the corresponding \gls{DSW} prediction model. However, even with the rather limited amount of reliability measures utilized in this study, the \gls{DSW}-\gls{EKF} yields a tracking performance that is similar to all baseline methods on this dataset. More advanced prediction models and the improved selection of reliability measures might help to shift the \gls{DSW}-\gls{EKF} tracking performance closer to the \gls{ODSW} limit.

Similar results were obtained for the \gls{NAVLoC} dataset. The \gls{DSW}-\gls{EKF} shows a statistically significant performance improvement compared to the \gls{PF}-based method proposed in~\cite{Gerlach2012}. There is only a marginal difference between the \gls{DSW}-\gls{EKF} and the \gls{ODSW}-\gls{EKF} and it performs better on average than the remaining baseline methods. It should be noted that this dataset is challenging especially for \gls{PF}-based methods, as the speaker position is fixed throughout all audiovisual sequences. This corresponds to a small process noise, which cannot be handled efficiently using \glspl{PF}. However, the \gls{PF} baseline method from~\cite{Qian2017} shows a performance comparable to the \gls{EKF}-based methods, which indicates that the \gls{MDF} approach used in this algorithm is efficient on this dataset.

Lastly, the proposed \gls{ODSW}-\gls{EKF} and \gls{DSW}-\gls{EKF} frameworks yield similar performance to the \gls{PF}-based methods on the \gls{MVAD} dataset. The average achieved azimuth errors indicate that the method from~\cite{Gerlach2012} outperforms the algorithms proposed in this study. However, due to the limited sample size, it is difficult to reliably show significant differences on this corpus. The fact that both \gls{DSW}-\gls{EKF} and \gls{ODSW}-\gls{EKF} algorithms have an identical average performance further indicates that the exploited reliability measures provide a suitable means to perform \gls{DSW} prediction on this dataset.

\subsection{Empirical analysis of computational complexity}
\label{subsec:computational_complexity}
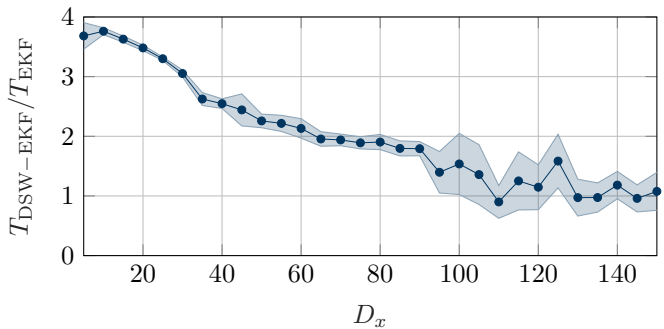
\begin{figure}[t]
	\centering
%
%
\definecolor{rub_blue}{rgb}{0.00000, 0.20784, 0.37647}%
\definecolor{rub_green}{rgb}{0.58039,0.75686,0.10588}%

\begin{tikzpicture}

\begin{axis}[%
width=3.0in,
height=1.25in,
at={(0in,0in)},
scale only axis,
xmin=5,
xmax=150,
xlabel style={font=\color{white!15!black}},
xlabel={\(D_{x}\)},
ymin=0,
ymax=4,
ylabel style={font=\color{white!15!black}},
ylabel={\(T_{\mathrm{DSW-EKF}} / T_{\mathrm{EKF}}\)},
y label style={yshift=-0.4cm},
axis background/.style={fill=white},
xmajorgrids,
ymajorgrids
]

\addplot[area legend, draw=none, fill=rub_blue, fill opacity=0.2]
table[row sep=crcr] {%
x	y\\
5	3.45954479464097\\
10	3.70441151426965\\
15	3.57625642946173\\
20	3.4343730805461\\
25	3.27074182455977\\
30	2.9944935040111\\
35	2.51669303511837\\
40	2.46670917987379\\
45	2.17412359668846\\
50	2.14358331799101\\
55	2.08145167396197\\
60	1.9693864952337\\
65	1.83323484393821\\
70	1.84165070612467\\
75	1.78839843781334\\
80	1.77657691192746\\
85	1.6707416213558\\
90	1.67365569635297\\
95	1.04966281729996\\
100	1.02328734068806\\
105	0.856879821611931\\
110	0.625294526359285\\
115	0.765226442914674\\
120	0.770099877397487\\
125	1.13559401552733\\
130	0.663559126303601\\
135	0.728918392175679\\
140	0.952645786485745\\
145	0.732422744933824\\
150	0.758895298909515\\
150	1.39442977197932\\
145	1.18590436920837\\
140	1.41072580610908\\
135	1.21832131382374\\
130	1.28017903792011\\
125	2.03338582961646\\
120	1.52271611193709\\
115	1.7389094035193\\
110	1.17465034538068\\
105	1.85935961760143\\
100	2.04943033451611\\
95	1.74413680989891\\
90	1.91105864476491\\
85	1.92319796051072\\
80	2.03116092120205\\
75	1.99233059924807\\
70	2.03649487341539\\
65	2.07663579266607\\
60	2.29467849158976\\
55	2.35171905240084\\
50	2.37225660798004\\
45	2.71036133383826\\
40	2.62760958567882\\
35	2.73389249410105\\
30	3.11017990577586\\
25	3.32976977848059\\
20	3.52820262958275\\
15	3.68104337045046\\
10	3.81905275789375\\
5	3.90563726834083\\
}--cycle;

\addplot [color=white!55!rub_blue]
  table[row sep=crcr]{%
5	3.45954479464097\\
10	3.70441151426965\\
15	3.57625642946173\\
20	3.4343730805461\\
25	3.27074182455977\\
30	2.9944935040111\\
35	2.51669303511837\\
40	2.46670917987379\\
45	2.17412359668846\\
50	2.14358331799101\\
55	2.08145167396197\\
60	1.9693864952337\\
65	1.83323484393821\\
70	1.84165070612467\\
75	1.78839843781334\\
80	1.77657691192746\\
85	1.6707416213558\\
90	1.67365569635297\\
95	1.04966281729996\\
100	1.02328734068806\\
105	0.856879821611931\\
110	0.625294526359285\\
115	0.765226442914674\\
120	0.770099877397487\\
125	1.13559401552733\\
130	0.663559126303601\\
135	0.728918392175679\\
140	0.952645786485745\\
145	0.732422744933824\\
150	0.758895298909515\\
};

\addplot [color=white!55!rub_blue]
  table[row sep=crcr]{%
5	3.90563726834083\\
10	3.81905275789375\\
15	3.68104337045046\\
20	3.52820262958275\\
25	3.32976977848059\\
30	3.11017990577586\\
35	2.73389249410105\\
40	2.62760958567882\\
45	2.71036133383826\\
50	2.37225660798004\\
55	2.35171905240084\\
60	2.29467849158976\\
65	2.07663579266607\\
70	2.03649487341539\\
75	1.99233059924807\\
80	2.03116092120205\\
85	1.92319796051072\\
90	1.91105864476491\\
95	1.74413680989891\\
100	2.04943033451611\\
105	1.85935961760143\\
110	1.17465034538068\\
115	1.7389094035193\\
120	1.52271611193709\\
125	2.03338582961646\\
130	1.28017903792011\\
135	1.21832131382374\\
140	1.41072580610908\\
145	1.18590436920837\\
150	1.39442977197932\\
};

\addplot [color=rub_blue, mark=*, line width=0.25, mark options={mark size=1.5pt, solid, fill=rub_blue, rub_blue}]
  table[row sep=crcr]{%
5	3.6825910314909\\
10	3.7617321360817\\
15	3.6286498999561\\
20	3.48128785506443\\
25	3.30025580152018\\
30	3.05233670489348\\
35	2.62529276460971\\
40	2.54715938277631\\
45	2.44224246526336\\
50	2.25791996298552\\
55	2.21658536318141\\
60	2.13203249341173\\
65	1.95493531830214\\
70	1.93907278977003\\
75	1.89036451853071\\
80	1.90386891656475\\
85	1.79696979093326\\
90	1.79235717055894\\
95	1.39689981359944\\
100	1.53635883760209\\
105	1.35811971960668\\
110	0.899972435869983\\
115	1.25206792321699\\
120	1.14640799466729\\
125	1.58448992257189\\
130	0.971869082111854\\
135	0.97361985299971\\
140	1.18168579629741\\
145	0.959163557071098\\
150	1.07666253544442\\
};

\end{axis}
\end{tikzpicture}%
	\caption{Mean and standard deviation of the computation time ratio between the standard \gls{EKF} and the proposed \gls{DSW}-\gls{EKF} for different state dimensionality. In all experiments, the observation dimension and number of of observations were set to \(D_{y_{m}} = 1\) and \(M = 2\), respectively.}
	\label{fig:complexity_state}
\end{figure}%

\begin{figure}[t]
	\centering
	\input{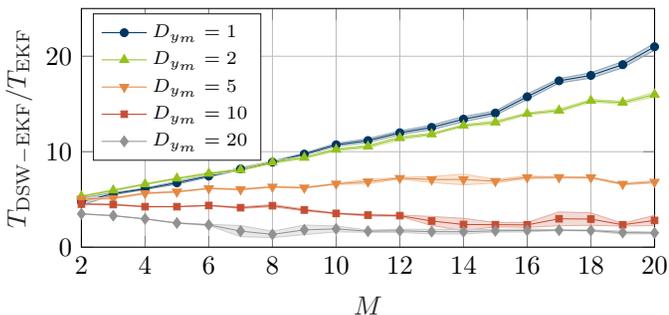}
	\caption{Mean and standard deviation of the computation time ratio between the standard \gls{EKF} and the proposed \gls{DSW}-\gls{EKF} for different observation dimensionality and number of independent observations. For the \gls{EKF}, varying numbers of observations were handled by stacking the individual observations within a single observation vector with dimensionality \(M D_{y_{m}}\). The state dimension was set to \(D_{x} = 1\) in all experiments.}
	\label{fig:complexity_observations}
\end{figure}%
The third evaluation scenario aims at empirically analyzing the computational performance of the proposed \gls{DSW}-\gls{EKF} compared to the standard \gls{EKF}. This analysis is conducted using synthetic data, generated from \glspl{DS} with varying state and observation dimensions. Figs.~\ref{fig:complexity_state} and \ref{fig:complexity_observations} depict the results of this analysis, which were obtained from Monte Carlo experiments with \(25\) runs per condition using randomly generated observation sequences with \(100\) time steps each. All model parameters were set to identity matrices if applicable, yielding linear models without the requirement of explicitly computing state and observation Jacobians. The experiments were conducted on a single desktop computer with an Intel\textsuperscript{\textregistered} Core\textsuperscript{\texttrademark} i5 processor and \(16\,\mathrm{GB}\) RAM running Ubuntu \(16.04\).

The results indicate that the standard \gls{EKF} is up to four times faster for simple models with low state and observation dimensionality. For increasing state dimension, this performance benefit vanishes and even decreases to similar computational performance for large state spaces with \(D_{x} > 100\). A similar effect is present for the observation dimensionality, where the \gls{EKF} outperforms the \gls{DSW}-\gls{EKF} for \(D_{y_{m}} \le 5\). However, with increasing observation dimensionality and number of independent observations, both \gls{EKF} and \gls{DSW}-\gls{EKF} show similar performance.

\section{Conclusion and Outlook}
\label{sec:conclusion_and_outlook}
In this study, a framework was introduced that extended the classical notion of dynamical systems with dynamic stream weights. A recursive state estimation scheme based on the Gaussian filtering paradigm was proposed. Additionally, a convex optimization approach to estimate oracle dynamic stream weights in fully observed dynamical systems utilizing a Dirichlet prior was derived. It was evaluated against a previously proposed method based on a Gaussian prior and the standard extended Kalman filter, showing similar performance with a reduced amount of tunable hyperparameters. A generic parameter learning framework for dynamic stream weight estimators was derived on the basis of previously computed oracle dynamic stream weights. A study using three different audiovisual speaker tracking datasets confirmed improved localization performance of the dynamic stream weight-based estimation framework over state-of-the-art methods.

Future research directions will focus on improving dynamic stream weight prediction models via suitable reliability measures. The measures utilized in this study only serve as a starting point and need to be investigated in depth and possibly be adapted to different applications. A thorough of these measures by means of feature selection may yield interesting theoretical insights towards the reliability of audiovisual sensors. Additionally, the extension to multi-speaker scenarios by incorporating probabilistic data association techniques into the tracking framework will make the proposed system suitable for a wider range of technical applications. A particular challenge in such scenarios will be the investigation of speaker-dependent reliability measures. Making the proposed system trainable end-to-end using deep neural networks might be a promising approach to tackle this particular challenge.

\appendices

\section{Solvability Analysis of the System of Linear Matrix Equations in Eq.~\eqref{eqn:sle}}
\label{sec:app:singularity_of_w}
The left-hand side of Eq.~\eqref{eqn:sle} contains the matrix expression \(\boldsymbol{R} + \boldsymbol{U}_{k} \boldsymbol{W}_{k}\boldsymbol{U}_{k}^{\mathrm{T}}\), which needs to be inverted to obtain a unique solution for the individual Kalman gains. As introduced in Sec.~\ref{subsec:state_estimation}, \(\boldsymbol{W}_{k} \in \mathbb{R}^{M D_{x} \times M D_{x}}\) can be expressed as the Kronecker product \(\boldsymbol{L}_{k} \otimes \hat{\boldsymbol{\Sigma}}_{k | k - 1}\), where \(\boldsymbol{L}_{k} \in \mathbb{R}^{M \times M}\) is given in Eq.~\eqref{eqn:lambda_matrix}. Since all rows in \(\boldsymbol{L}_{k}\) are linearly dependent, \(\mathrm{rank}(\boldsymbol{L}_{k}) = 1\). Additionally, \(\hat{\boldsymbol{\Sigma}}_{k | k - 1}\) is a \(D_{x}\)-dimensional covariance matrix with \(\mathrm{rank}(\hat{\boldsymbol{\Sigma}}_{k | k - 1}) = D_{x}\). Hence, the rank equality of the Kronecker product~\cite{Tian2005} can be exploited here, which yields \(\mathrm{rank}(\boldsymbol{W}_{k}) = \mathrm{rank}(\boldsymbol{L}_{k}) \cdot~\mathrm{rank}(\hat{\boldsymbol{\Sigma}}_{k | k - 1}) = D_{x}\). Therefore, \(\boldsymbol{W}_{k}\) is singular for \(M > 1\), which requires the appropriate form of the binomial inverse theorem~\cite{Harville1976} to obtain a unique solution for Eq.~\eqref{eqn:sle}.

\section{Proof of Concavity of the Fully-Observed Likelihood function with Dirichlet prior}
\label{sec:app:concave}
The first and second derivatives of the log-likelihood function defined in Eq.~\eqref{eqn:log_likelihood_with_dirichlet_prior} are
\begin{equation*}
\frac{\partial}{\partial\lambda_{m, k}}\,J(\boldsymbol{\lambda}_{k}) = \log\{p(\boldsymbol{y}_{m, k}\,|\,\boldsymbol{x}_{k})\} + \frac{\alpha - 1}{\lambda_{m, k}}
\end{equation*}
and
\begin{equation*}
\frac{\partial^{2}}{\partial\lambda_{m, k}^{2}}\, J(\boldsymbol{\lambda}_{k}) = \frac{1 - \alpha}{\lambda_{m, k}^{2}}.
\end{equation*}	
The second derivative is negative with the constraint \(\alpha > 1\) for \(0 < \lambda_{m, k} < 1~\forall\,m\). Hence, the first derivative is a strictly monotonically decreasing function in this parameter range, which implies that the log-likelihood function is strictly concave, cf.~\cite[Chap.~3]{Boyd2004}.

\ifCLASSOPTIONcaptionsoff
  \newpage
\fi



\bibliographystyle{IEEEtran}
\bibliography{references}

\vfill

\end{document}